\documentclass[lettersize,journal]{IEEEtran}
\usepackage{amsmath,amsfonts}
\usepackage{algorithmic}
\usepackage{algorithm}
\usepackage{array}
\usepackage{textcomp}
\usepackage{stfloats}
\usepackage{url}
\usepackage{verbatim}
\usepackage{graphicx}
\usepackage{cite}

\usepackage{arydshln} 

\usepackage{color}
\usepackage{xcolor}

\usepackage{booktabs}
\usepackage{subfigure}
\usepackage{bm,multirow,tablefootnote}
\usepackage{threeparttable}
\usepackage{bbding}

\definecolor{cvprblue}{rgb}{0.21,0.49,0.74}
\usepackage[breaklinks,colorlinks,citecolor=cvprblue]{hyperref}
\begin{document}

\title{Deep Learning for Video Anomaly Detection:\\ A Review}

\author{Peng Wu, Chengyu Pan, Yuting Yan, Guansong Pang, Peng Wang,  and \\Yanning Zhang~\IEEEmembership{Senior Member, IEEE}
\thanks{Peng Wu, Chengyu Pan, Yuting Yan, Peng Wang, and Yanning Zhang are with the School of Computer Science, Northwestern Polytechnical University, China. E-mail: xdwupeng@gmail.com;\{peng.wang, ynzhang\}@nwpu.edu.cn.

Guansong Pang is with the School of Computing and Information Systems, Singapore Management University Singapore, Singapore. E-mail: pangguansong@gmail.com.} 

\thanks{Manuscript received April 19, 2021; revised August 16, 2021. (Corresponding author: Guansong Pang. Chengyu Pan and Yuting Yan contributed equally.)}}

\markboth{Journal of \LaTeX\ Class Files,~Vol.~14, No.~8, August~2021}%
{Shell \MakeLowercase{\textit{et al.}}: A Sample Article Using IEEEtran.cls for IEEE Journals}


\maketitle
\begin{abstract}
Video anomaly detection (VAD) aims to discover behaviors or events deviating from the normality in videos. As a long-standing task in the field of computer vision, VAD has witnessed much good progress. In the era of deep learning, with the explosion of architectures of continuously growing capability and capacity, a great variety of deep learning based methods are constantly emerging for the VAD task, greatly improving the generalization ability of detection algorithms and broadening the application scenarios. Therefore, such a multitude of methods and a large body of literature make a comprehensive survey a pressing necessity. In this paper, we present an extensive and comprehensive research review, covering the spectrum of five different categories, namely, semi-supervised, weakly supervised, fully supervised, unsupervised and open-set supervised VAD, and we also delve into the latest VAD works based on pre-trained large models, remedying the limitations of past reviews in terms of only focusing on semi-supervised VAD and small model based methods. For the VAD task with different levels of supervision, we construct a well-organized taxonomy, profoundly discuss the characteristics of different types of methods, and show their performance comparisons. In addition, this review involves the public datasets, open-source codes, and evaluation metrics covering all the aforementioned VAD tasks. Finally, we provide several important research directions for the VAD community. 
 
\end{abstract}


\section{Introduction}
\IEEEPARstart{A}{nomaly} represents something that deviates from what is standard, normal, or expected. There are myriads of normalities, and anomalies are considerably scarce. However, when anomalies do appear, they often have a negative impact. Anomaly detection aims to discover these rare anomalies built on top of machine learning, thereby reducing the cost of manual judgment. Anomaly detection has widespread application across various fields\cite{pang2021deep}, such as financial fraud detection, network intrusion detection, industrial defect detection, and human violence detection. Among these, video anomaly detection (VAD) occupies an important place, in which anomaly indicates the abnormal events in the temporal or spatial dimensions. VAD not only plays a vital role in intelligent security (e.g., violence, intrusion, and loitering detection) but is also widely used in other scenarios, such as online video content review and traffic anomaly prediction in autonomous driving\cite{yao2022dota}. Owing to its significant potential for applications across different fields, VAD has attracted considerable attention from both industry and academia.

\begin{figure}[t]
  \centering
  \includegraphics[width=0.9\linewidth]{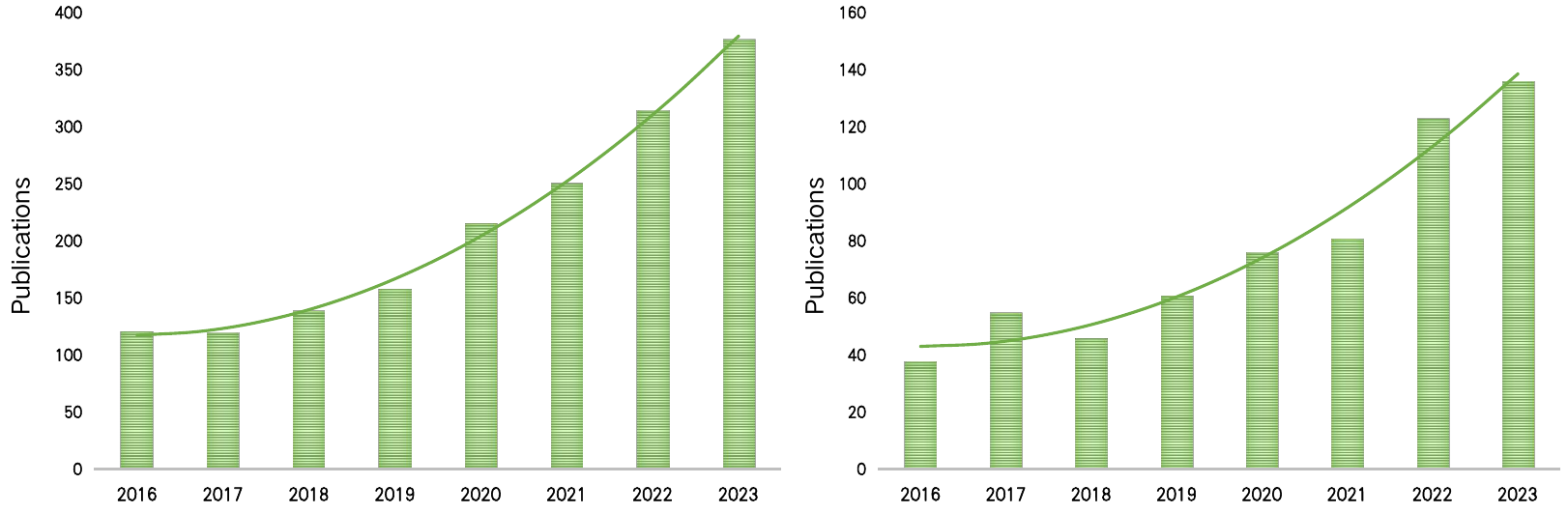}
  \caption{Publications on VAD. Left: Google Scholar; Right: IEEE Xplore.}
  \label{pub}
  \vspace{-0.2cm} 
\end{figure}

\begin{figure}[t]
  \centering
  \includegraphics[width=0.9\linewidth]{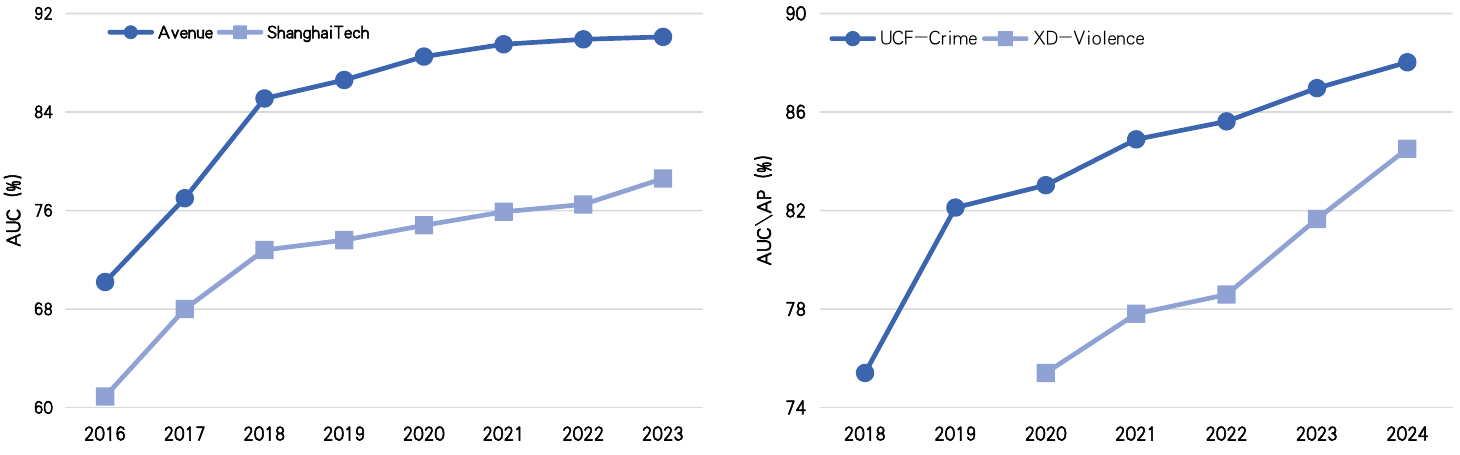}
  \caption{Performance development for semi/weakly supervised VAD tasks.}
  \label{performance}
  \vspace{-0.6cm} 
\end{figure}

In the pre-deep learning era, the routine way is to separate feature extraction and classifier design, which forms a two-stage process, and then combine them together during the inference stage. First, there is a feature extraction process to convert the original high dimensional raw videos into compact hand-crafted features based on prior knowledge of experts. 
Although hand-crafted features lack robustness and are difficult to use for capturing effective behavior expressions in the face of complex scenarios,
these pioneer works deeply enlighten subsequent deep learning based works.

The rise of deep learning has made traditional machine learning algorithms fall out of favor over the last decade. With the rapid development of computer hardware and the massive data in the Internet era, we have witnessed great progress in developing deep learning based methods for VAD in recent years. For example, ConvAE\cite{hasan2016learning}, the first work using deep auto-encoders based on convolutional neural networks (CNNs) for capturing the regularities in videos; FuturePred\cite{liu2018future}, the first work making use of U-Net for forecasting future anomalies; DeepMIL\cite{sultani2018real}, the first endeavor exploring deep multiple instance learning (MIL) framework for real-world anomalies. In order to more intuitively manifest the research enthusiasm for the VAD task in the era of deep learning, we conduct a statistical survey on the number of publications related to VAD over the past decade (the era driven by the rise of deep learning based methods) through Google Scholar and IEEE Xplore\footnote{\href{https://scholar.google.com}{https://scholar.google.com}; \href{https://ieeexplore.ieee.org/}{https://ieeexplore.ieee.org/}}. We select five related topics, i.e., video anomaly detection, abnormal event detection, abnormal behavior detection, anomalous event detection, and anomalous behavior detection, and showcase the publication statistics in Figure~\ref{pub}. It is not hard to see that the number of related publications counted from both sources exhibits a steady and rapid growth trend, demonstrating that VAD has garnered widespread attention. Moreover, we also demonstrate the detection performance trends of annual state-of-the-art methods on commonly used datasets under two common supervised manners, and present performance trends in Figure~\ref{performance}. The detection performance shows a steady upward trend across all datasets, without displaying any performance bottleneck. For instance, the performance of semi-supervised methods on CUHK Avenue\cite{lu2013abnormal} has experienced a significant surge, rising from 70.2\% AUC\cite{hasan2016learning} to an impressive 90.1\% AUC\cite{yan2023feature} over the past seven-year period. Moreover, for the subsequently proposed weakly supervised VAD, significant progress has been achieved as well. 
This indicates the evolving capability of deep learning methods under developing architectures, and also showcases the ongoing exploration enthusiasm of deep learning methods for the VAD task.

\begin{table*}[]
  \centering
  \caption{Analysis and Comparison of Related Reviews.}
  \label{tab:survey}
  \resizebox{\textwidth}{!}{
  \begin{tabular}{lccc|ccccc:cc}
  \toprule
  \multirow{2}{*}{\textbf{Reference}} &\multirow{2}{*}{\textbf{Year}} & \multirow{2}{*}{\textbf{Main Focus}}        & \multicolumn{1}{c}{\multirow{2}{*}{\textbf{Main Categorization}}} & \multicolumn{7}{c}{\textbf{Research Topics}}  \\ \cline{5-11} 
  &    &   & \multicolumn{1}{l}{}    & UVAD      & WVAD      & SVAD      & FVAD & OVAD& LVAD & IVAD\\ \hline
  Ramachandra et al.\cite{ramachandra2020survey}  &2020     &  Semi-supervised single-scene VAD   & Methodology    & \footnotesize\XSolidBrush   & \footnotesize\XSolidBrush & \footnotesize\Checkmark   & \footnotesize\XSolidBrush & \footnotesize\XSolidBrush& \footnotesize\XSolidBrush & \footnotesize\XSolidBrush  \\
  Santhosh et al.\cite{santhosh2020anomaly}    &2020      &  VAD applied on road traffic  & Methodology & \footnotesize\Checkmark   & \footnotesize\XSolidBrush & \footnotesize\Checkmark & \footnotesize\Checkmark & \footnotesize\XSolidBrush & \footnotesize\XSolidBrush & \footnotesize\XSolidBrush  \\
  
  Nayak et al.\cite{nayak2021comprehensive} & 2021     &Deep learning driven semi-supervised VAD  & Methodology & \footnotesize\XSolidBrush   & \footnotesize\XSolidBrush & \footnotesize\Checkmark & \footnotesize\XSolidBrush & \footnotesize\XSolidBrush& \footnotesize\XSolidBrush & \footnotesize\XSolidBrush   \\ 
   
  Tran et al.\cite{tran2022anomaly}  &2022 &  Semi\&weakly supervised VAD &  Architecture &\footnotesize\XSolidBrush   & \footnotesize\XSolidBrush   & \footnotesize\Checkmark   & \footnotesize\XSolidBrush  & \footnotesize\XSolidBrush& \footnotesize\XSolidBrush & \footnotesize\XSolidBrush  \\ 
  
  Chandrakala et al.\cite{chandrakala2023anomaly}   &2023   &   Deep model-based one\&two-class VAD & Methodology\&Architecture   & \footnotesize\XSolidBrush   & \footnotesize\Checkmark   & \footnotesize\Checkmark   & \footnotesize\Checkmark  & \footnotesize\XSolidBrush & \footnotesize\XSolidBrush & \footnotesize\XSolidBrush \\ 
  
  Liu et al.\cite{liu2023generalized}   & 2023   & Deep models for semi\&weakly supervised VAD  & Model Input  & \footnotesize\Checkmark   & \footnotesize\Checkmark   & \footnotesize\Checkmark   & \footnotesize\Checkmark  & \footnotesize\XSolidBrush & \footnotesize\XSolidBrush & \footnotesize\XSolidBrush \\ \hline
  
   \multirow{2}{*}{Our survey}  & \multirow{2}{*}{2024}   & \multirow{2}{*}{Comprehensive VAD taxonomy and deep models}  & Methodology, Architecture, Refinement&\multirow{2}{*}{\footnotesize\Checkmark}   & \multirow{2}{*}{\footnotesize\Checkmark}   & \multirow{2}{*}{\footnotesize\Checkmark}  & \multirow{2}{*}{\footnotesize\Checkmark} & \multirow{2}{*}{\footnotesize\Checkmark} & \multirow{2}{*}{\footnotesize\Checkmark} & \multirow{2}{*}{\footnotesize\Checkmark} \\ 
   ~& ~ &~ &Model Input, Model Output & ~& ~& ~& ~& ~& ~& ~\\
   \bottomrule
  \end{tabular}}
  \begin{tablenotes}    
     \footnotesize   
     \item UVAD: Unsupervised VAD, WVAD: Weakly supervised VAD, SVAD: Semi-supervised VAD, FVAD: Fully supervised VAD, OVAD: Open-set supervised VAD, LVAD: Large-model based VAD, IVAD: Interpretable VAD
  \end{tablenotes}   
  \vspace{-0.4cm} 
\end{table*}

The above statistics clearly demonstrate that deep learning driven VAD is the hot area of the current research. Therefore, there is an urgent necessity for a systematic taxonomy and comprehensive summary of existing works, to facilitate newcomers as a guide and provide references for existing researchers. Based on this, we first collect some high-profile reviews on VAD in the past few years, which are shown in Table~\ref{tab:survey}. Ramachandra et al.\cite{ramachandra2020survey} mainly focused on semi-supervised VAD in the single scenario, lacking in discussions of cross scenes. Santhosh et al.\cite{santhosh2020anomaly} reviewed VAD methods focusing on entities in road traffic scenarios. Their reviews lack sufficient in-depth analysis and center on pre-2020 methodologies, resulting in the neglect of recent advances. Nayak et al.\cite{nayak2021comprehensive} comprehensively surveyed on deep learning based methods for semi-supervised VAD, but did not take into account weakly supervised VAD methods. The follow-up work Tran et al.\cite{tran2022anomaly} introduced a review of the emerging weakly supervised VAD, but the focus is not only on videos but also on image anomaly detection, resulting in a lack of systematic organization of the VAD task. More recently, both Chandrakala et al.\cite{chandrakala2023anomaly} and Liu et al.\cite{liu2023generalized} constructed an organized taxonomy covering a variety of VAD tasks, e.g., unsupervised VAD, semi-supervised VAD, weakly supervised VAD, and fully supervised VAD, and also surveyed deep learning based methods for most supervised VAD tasks. However, they restrict their scope to the conventional close-set scenario, and fail to cover the latest research in the field of open-set supervised VAD, without introducing a brand-new pipeline based on pre-trained large models and interpretable learning.

To address this gap comprehensively, we present a thorough survey of VAD works in the deep learning era. Our survey covers several key aspects to provide a comprehensive analysis of VAD studies. To be specific, we perform an in-depth investigation into the development trends of VAD task in the era of deep learning, and then propose a unified framework that integrates different VAD tasks together, filling the gaps in the existing reviews in terms of taxonomy. We then collect the most comprehensive open sources, including benchmark datasets, evaluation metrics, open-source codes, and performance comparisons, to help researchers in this field avoid detours and improve efficiency. Further, we systematically categorize various VAD tasks, dividing existing works into different categories and establishing a clear and structured taxonomy system to provide a coherent and organized overview of various VAD paradigms. In addition to this taxonomy, we conduct a comprehensive analysis of each paradigm. Furthermore, throughout this survey, we spotlight influential works that have significantly contributed to the research advancement in VAD.

The main contributions of this survey are summarized in the following three aspects:
\begin{itemize}
\item 
We provide a comprehensive review of VAD, covering five tasks based on different supervision signals, i.e., semi-supervised VAD, weakly supervised VAD, fully supervised VAD, unsupervised VAD, and open-set supervised VAD. The research focus has expanded from traditional single-task semi-supervised VAD to a broader range of multiple VAD tasks.
\item Staying abreast of the research trends, we review the latest studies on open-set supervised VAD. Moreover, we also revisit the most recent VAD methods based on pre-trained large models and interpretable learning. The emergence of these methods elevates both the performance and application prospects of VAD. To our knowledge, this is the first comprehensive survey of open-set supervised VAD and pre-trained large model based VAD methods.
\item For different tasks, we systematically review existing deep learning based methods, and more importantly, we introduce a unified taxonomy framework categorizing the methods from various VAD paradigms based on various aspects, including model input, architecture, methodology, model refinement, and output. This meticulous scientific taxonomy enables a comprehensive understanding of the field.
\end{itemize}

\section{Background}\label{back}

\subsection{Notation and Taxonomy}
\begin{figure*}[t]
  \centering
  \includegraphics[width=0.8\linewidth]{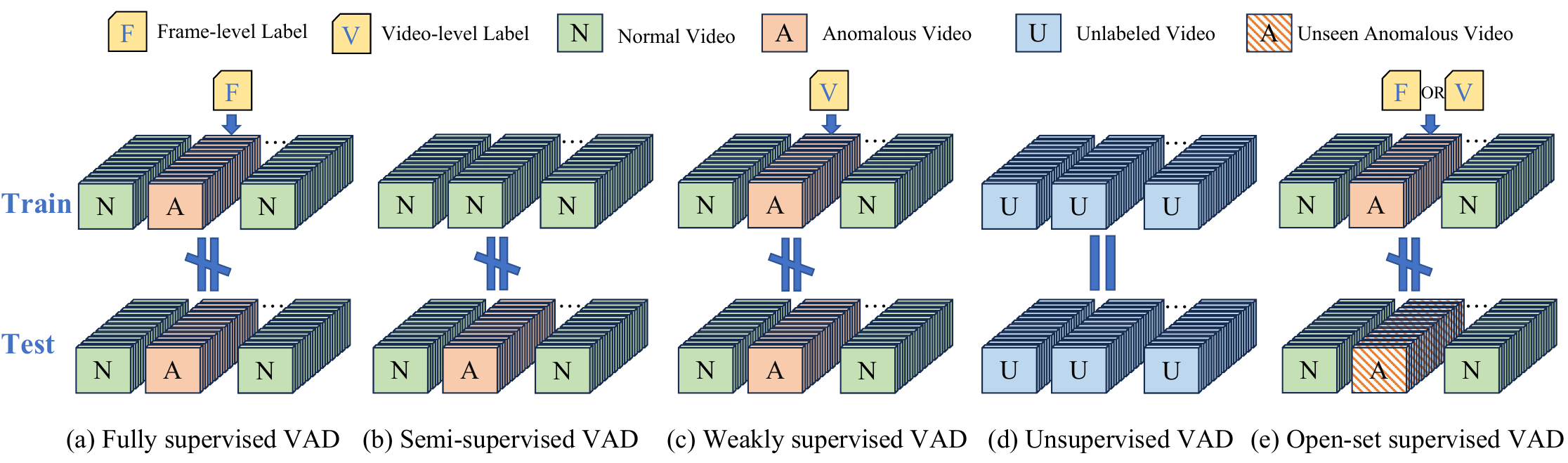}
  \caption{Comparisons of five supervised VAD tasks, i.e., fully supervised, semi-supervised, weakly supervised, unsupervised, and open-set supervised VAD.}
  \label{fig:vads}
  \vspace{-0.3cm} 
\end{figure*}

As aforementioned, the studied problem, VAD, can be formally divided into five categories based on supervision signals. Different supervised VAD tasks aim to identify anomalous behaviors or events, but with different training and testing setups. We demonstrate these different VAD tasks in Figure~\ref{fig:vads}.

The general VAD problem is presented as follows. Suppose we are given a set of training samples $\mathcal{X}=\{x_i\}_{i=1}^{N+A}$ and corresponding labels $\mathcal{Y}$, where $\mathcal{X}_n=\{x_i\}_{i=1}^{N}$ is the set of normal samples and $\mathcal{X}_a=\{x_i\}_{i=N+1}^{N+A}$ is the set of abnormal samples. Each sample $x_i$ is accompanied by a corresponding supervision label $y_i$ in $\mathcal{Y}$. During the training phase, the detection model $\Phi(\theta)$ takes $\mathcal{X}$ as input and generates anomaly predictions; it is then optimized according to the following objective,
\begin{equation}
 l= \mathcal{L}\left(\Phi\left(\theta, \mathcal{X}\right), \mathcal{X} \; or \;\mathcal{Y}\right)
\label{}
\end{equation}
where $\mathcal{L}(\cdot)$ is employed to quantify the discrepancy between the predictions and the ground-truth labels or original samples. 
During inference, the detection model is expected to locate the abnormal behaviors or events in videos based on the generated anomaly predictions. 
Depending on the input to $\mathcal{L}$, VAD can be categorized into one of the following five task settings.

\textbf{Semi-supervised VAD} assumes that only normal samples are available during the training stage, that is, $\mathcal{X}_a=\emptyset$. This task aims to learn the normal patterns based on the training samples and consider the test samples which fall outside the learned patterns as anomalies. \textit{Pros and Cons:} Only normal samples are required for training, hence, there is no need to painstakingly collect scarce abnormal samples. However, any unseen test sample may be recognized as abnormal, leading to a higher false positive rate.

\textbf{Weakly supervised VAD} has more sufficient training samples and supervision signals than semi-supervised VAD. Both normal and abnormal samples are provided during the training stage, but the precise location stamps of anomalies in these untrimmed videos are unknown. In other words, only coarse video-level labels are available (i.e., inexact supervision). Formally, $\mathcal{Y}=\{0, 1\}_{i=1}^{N+A}$, where $y_i=0$ indicates that $x_i$ is normal, and $y_i=1$ indicates that $x_i$ is abnormal. \textit{Pros and Cons:} Compared to fully supervised annotations, it can significantly reduce labeling costs. However, it places higher demands on algorithm design and may lead to situations of blind guessing.

\textbf{Fully supervised VAD}, as its name implies, comprises the complete supervision signals, meaning that each abnormal sample has precise annotations of anomalies. This task can be viewed as a standard video or frame classification problem. Due to the scarcity of abnormal behaviors and intensive manual labeling in reality, there has been little research on the fully supervised VAD task. It is noteworthy that video violence detection can be regarded as a fully supervised VAD, hence, we denote violence detection as a fully supervised VAD task in this paper. Formally, each video $x_i$ in $\mathcal{X}_a$ is accompanied by a corresponding supervision label $y_i=\{(t^s_j, t^e_j)\}^{U_i}_j$, where $t^s_j$ and $t^e_j$ denote the start and end time of the $j$-th violence event, $U_i$ indicates the total number of anomalies present in the video. \textit{Pros and Cons:} In contrast to weakly supervised VAD, with full supervised information, the detection performance of the algorithms would be remarkable. However, the corresponding drawback is the high requirement for intensive manual annotations.

\textbf{Unsupervised VAD} aims to discover anomalies directly from fully unlabeled videos in an unsupervised manner. Thus, unsupervised VAD no longer requires labeling normal and abnormal videos to build the training set. It can be expressed formally as follows, $\mathcal{X}=\mathcal{X}_{test}$, and $\mathcal{Y}=\emptyset$, in which $\mathcal{X}_{test}$ denotes the set of test samples. \textit{Pros and Cons:} No time-consuming effort is needed to collect training samples, avoiding the heavy labeling burden. Besides, this assumption also expands the application fields of VAD, implying that the detection system can continuously retrain without human intervention. Unfortunately, due to the lack of labels, the detection performance is relatively poor, leading to a higher rate of false positives and false negatives.

\textbf{Open-set supervised VAD} is devised to discover unseen anomalies that are not presented in the training set. Unlike semi-supervised VAD, open-set supervised VAD includes abnormal samples in the training set, which are referred to as seen anomalies. 
Specifically, for each $x_i$ in $\mathcal{X}_a$, its corresponding label $y_i \in C_{base}$, here $C_{base}$ represents the set of base (seen) anomaly categories, and $C_{base} \subset \mathcal{C}$, with $\mathcal{C}=C_{base}\cup C_{novel}$. Here, $C_{novel}$ and $\mathcal{C}$ represent the sets of novel anomaly categories unseen during training and all anomaly categories, respectively. Given a testing sample $x_{test}$, its label $y_{test}$ may be either $\in C_{base}$ or $\in C_{novel}$. \textit{Pros and Cons:} Compared to the two most common tasks, i.e., semi-supervised VAD and weakly supervised VAD, open-set supervised VAD not only reduces false positives but also avoids being limited to closed-set scenarios, thus demonstrating high practical value. However, it relies on learning specialized classifiers, loss functions, or generating unknown classes to to detect unseen anomalies.

\subsection{Datasets and Metrics}
Related benchmark datasets and evaluation metrics are listed at \href{https://roc-ng.github.io/DeepVAD/}{https://roc-ng.github.io/DeepVAD/}.

\section{Semi-supervised Video Anomaly Detection}\label{semi}

\begin{figure*}[t]
  \centering
  \includegraphics[width=0.8\linewidth]{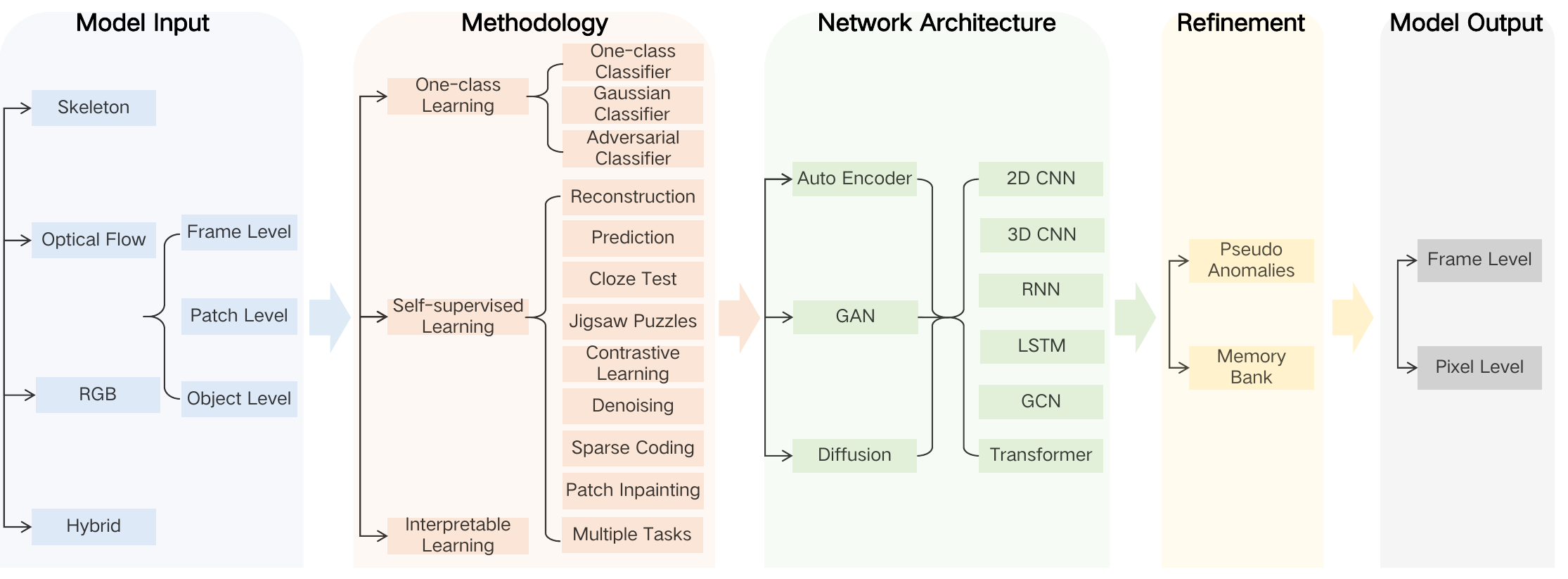}
\caption{The taxonomy of semi-supervised VAD. We provide a hierarchical taxonomy that organizes existing deep semi-supervised VAD models by model input, methodology, network architecture, refinement strategy, and model output into a systematic framework.}
  \label{fig: semiframework}
  \vspace{-0.4cm} 
\end{figure*}

Based on our in-depth investigation of past surveys, we found that previous surveys mostly lack a scientific taxonomy, in which many surveys simply categorize semi-supervised VAD works into different groups based on usage approaches, such as reconstruction-based, distance-based, and probability-based approaches, and some surveys classify works according to inputs, such as image-based, optical flow-based, and patch-based approaches. It is particularly apparent that existing classification reviews are relatively simplistic and superficial, thus making it challenging to cover all methods comprehensively and effectively. To address this issue, we establish a comprehensive taxonomy, encompassing model input, methodology, architecture, model refinement, and model output. The detailed illustration is presented in Figure~\ref{fig: semiframework}.

As aforementioned, only normal samples are available for training in the semi-supervised VAD task, rendering the supervised classification paradigm inapplicable. Common approaches involve leveraging the intrinsic information of the training samples to learn deep neural networks (DNNs) for solving a pretext task. For instance, normality reconstruction is a classic pretext task\cite{hasan2016learning}. During this process, several critical aspects need consideration: selection of sample information (Model Input), design of pretext tasks (Methodology), utilization of deep networks (Network Architecture), improvement of methods (Refinement), and expression of anomaly results (Model Output). 
These key elements collectively contribute to the effectiveness of semi-supervised VAD solutions. In the following sections, we introduce existing deep learning based VAD methods systematically according to the aforementioned taxonomy.

\subsection{Model Input}
Existing semi-supervised VAD methods typically use the raw video or its intuitive representations as the model input. Depending on the modality, these can be categorized as follows: RGB images, optical flow, skeleton, and hybrid inputs, where the first three represent appearance, motion, and body posture, respectively.

\subsubsection{RGB}
RGB images are the most common input for conventional vision tasks driven by deep learning techniques, and this holds true for the VAD task as well. Unlike other modalities, RGB images do not require additional processing steps such as optical flow calculations or pose estimation algorithms. In deep learning era, various deep models can be employed to extract compact and high-level visual features from these high-dimensional raw data. Utilizing these high-level features enables the design of more effective subsequent detection methods. Moreover, depending on the input size, RGB image based input can be categorized into three principal groups: frame level, patch level, and object level.

\textbf{Frame-level RGB} input provides a macroscopic view of the entire scene, encompassing both the background, which is usually unrelated to the event, and the foreground objects, where anomalies are more likely to occur. The conventional approach typically uses multiple consecutive video frames as a single input to capture temporal context within the video, as seen in methods like ConvAE\cite{hasan2016learning}, ConvLSTM-AE\cite{luo2017remembering}, and STAE\cite{zhao2017spatio}. On the other hand, several research studies focus on using single-frame RGB as input, aiming to detect anomalies at the spatial level, such as AnomalyGAN\cite{ravanbakhsh2017abnormal} and AMC\cite{nguyen2019anomaly}.

\textbf{Patch-level RGB} input involves segmenting the frame-level RGB input spatially or spatio-temporally, which focuses on local regions, effectively separating the foreground from the background and differentiating between various individual entities. The primary advantage of patch-level input is its ability to significantly reduce the interference from the dominant background, which is usually less relevant to the anomalies. This segmentation helps in isolating areas that are more likely to contain anomalies, thereby enhancing the detection accuracy. For example, AMDN\cite{xu2015learning, xu2017detecting}, DeepOC\cite{wu2019deep} and Deep-cascade\cite{sabokrou2017deep} took spatio-temporal patches as the input, and S$^2$-VAE\cite{wang2018generative} and GM-VAE\cite{fan2020video} only employed an image patch from a single video frame as the model input.

\textbf{Object-level RGB} input has emerged in recent years alongside the advancement of object detection approaches, which focuses exclusively on foreground objects. Compared to patch-level input, it entirely disregards background information and neglects to consider the relationship between objects and backgrounds. As a result, it demonstrates impressive performance in identifying anomalous events within complex scenes.  Hinami et al.\cite{hinami2017joint} first proposed an object-centric approach FRCN based on object inputs, a follow-up work ObjectAE\cite{ionescu2019object} introduced to train object-centric auto-encoders on detected objects, and subsequently, a series of works focusing on object-level inputs emerged, e.g., HF$^2$-VAD\cite{liu2021hybrid}, HSNBM\cite{bao2022hierarchical}, BDPN\cite{chen2022comprehensive}, ER-VAD\cite{sun2022evidential}, and HSC\cite{sun2023hierarchical}.

\subsubsection{Optical Flow}
A video is not merely a sequence of stacked RGB images, it encapsulates the temporal dimension and crucial temporal context. Therefore, extracting temporal context is vital for understanding video content, with motion information playing an irreplaceable role. Optical flow represents the motion information between consecutive video frames and is commonly used as a model input for VAD task. It typically does not appear alone but is paired with the corresponding RGB image as input in the two-stream network. Therefore, it also encompasses frame\cite{liu2018future, luo2021future, wu2020fast, cai2021appearance, liu2022learning, huang2023video}, patch\cite{wu2019deep, li2020spatial, ramachandra2020learning}, and object\cite{liu2021hybrid, sun2022evidential, reiss2022attribute, huang2023video} levels. 

\subsubsection{Skeleton}

In recent years, with the remarkable success of deep learning technologies in the field of pose estimation, VAD methods based on skeleton input have emerged. Skeleton input solely focuses on the human body, making it more specialized than object-level RGB input. It demonstrates impressive performance in human-centric VAD, marking it as a significant research interest in recent years within the VAD domain. Morais et al.\cite{morais2019learning} first endeavored to learn the normal patterns of human movements using dynamic skeleton, where the pose estimation is utilized to independently detect skeletons in each video frame. Then, GEPC\cite{markovitz2020graph}, MTTP\cite{rodrigues2020multi}, NormalGraph\cite{luo2021normal}, HSTGCNN\cite{zeng2021hierarchical}, TSIF\cite{yang2022two}, STGCAE-LSTM\cite{li2022human}, STGformer\cite{huang2022hierarchical}, STG-NF\cite{hirschorn2023normalizing}, MoPRL\cite{yu2023regularity}, MoCoDAD\cite{flaborea2023multimodal}, and TrajREC\cite{stergiou2024holistic} are specialized in human-related VAD with the skeleton input. 

\subsubsection{Hybrid}

The hybrid input from different modalities often proves more advantageous for the VAD task compared to the unimodal input due to their complementary nature. In existing deep learning driven VAD methods, hybrid input is a common practice. Typical hybrid input includes frame-level RGB combined with optical flow\cite{liu2018future}, patch-level RGB combined with optical flow\cite{wu2019deep}, and object-level RGB combined with optical flow\cite{liu2021hybrid}. More recently, several research studies have explored hybrid input based on RGB combined with skeleton\cite{pi2024eogt}.

\subsection{Methodology}
For semi-supervised VAD, only normal samples are provided during the training phase, rendering conventional supervised classification methods inapplicable. The current approach involves designing a pretext task based on the properties inherent in normal samples themselves to encapsulate a paradigm that encompasses all normal events, referred to as the normal paradigm or normal pattern. Through extensive research on existing works, we categorize three major approaches for learning the normal paradigm: self-supervised learning, one-class learning, and interpretable learning.

\subsubsection{Self-supervised Learning}

\begin{quotation}
``\textit{If intelligence is a cake, the bulk of the cake is self-supervised learning.}'' ~~~~~~~~~------~~\textit{Yann LeCun} 
\end{quotation}

Self-supervised learning primarily leverages auxiliary tasks (pretext tasks) to derive supervisory signals directly from unsupervised data. Essentially, self-supervised learning operates without external labeled data, as these labels are generated from the input data itself. For semi-supervised VAD task, which lacks explicit supervisory signals, self-supervised learning naturally becomes essential for learning normal representations and constructing the normal paradigm based on these auxiliary tasks. Consequently, self-supervised learning based methods consistently dominate the leading position in semi-supervised VAD task. Throughout this process, a significant research focus and challenge lie in designing effective pretext tasks derived from the data itself. Here, we compile the common design principles of auxiliary tasks used in existing self-supervised learning based methods.

\textbf{Reconstruction} is currently the most commonly used pretext task for self-supervised learning based methods in the field of semi-supervised VAD\cite{hasan2016learning, luo2017remembering, nguyen2019anomaly, wang2018generative, chang2020clustering, fang2020anomaly, huang2021self, yu2023regularity}. The main process involves inputting normal data into the network, performing encoding-decoding operations, and generating reconstructed data, encouraging the network to produce reconstructions that closely match the original input data. The objective can be expressed as,
\begin{equation}
 l_{rec}= \mathcal{L}\left(\Phi\left(\theta, x\right), x\right)
\label{rec}
\end{equation}
For convenience, in the following sections, unless otherwise specified, $x$ represent normal data, which could be a normal video, a normal video frame, a normal feature, or similar. The above objective function measures the reconstruction error, which serves as a criterion for determining whether the test data is anomalous during the test stage. The larger the reconstruction error, the higher the probability that the data is considered anomalous.
However, due to the high capacity of deep neural networks, reconstruction based methods cannot guarantee that larger reconstruction errors for abnormal events do necessarily happen.

\textbf{Prediction} fully leverages the temporal coherence inherent in videos, which is also a commonly used pretext task. This pretext task is based on the assumption that normal events are predictable, while abnormal ones are unpredictable. Specifically, the prediction pretext task takes historical data as input and, through encoding-decoding operations within the network, outputs the predicted data for the current moment. The network is compelled to make the predicted data similar to the actual current data. We define the optimization objective for prediction as,
\begin{equation}
 l_{pre}= \mathcal{L}\left(\Phi\left(\theta, \{I_{t-\Delta t}, ..., I_{t-1}\}\right), I_{t}\right)
\label{}
\end{equation}
${I}_{t}$ is the actual data at the current time step $t$, ${I}_{t-\Delta t:t-1}$ represents the historical data from time step $t-\Delta t$ to $t-1$. FuturePred\cite{liu2018future}, as a future frame prediction framework, provided a new solution for VAD. Then, many researchers \cite{zhou2019attention, zhang2020normality, cai2021appearance, liu2021hybrid, wang2021robust, yu2021abnormal, huang2022hierarchical, zhou2022object, cheng2023spatial, liu2023amp} proposed other prediction based methods. It alleviates, to some extent, the problem in reconstruction based methods where abnormal events can also be well reconstructed. 

\textbf{Visual cloze test} is inspired by the cloze test in natural language processing\cite{yu2020cloze, yang2023video, yu2023video}. It mainly involves training multiple DNNs to infer deliberately erased data from incomplete video sequences, where the prediction task can be considered a special case of visual cloze test task, i.e., the erased data happens to be the last frame in the video sequence. We define the objective function for completing erased data at the $t$-th time stamp as,
\begin{equation}
 l_{vct}^{(t)}= \mathcal{L}\left(\Phi\left(\theta, \{I_{1}, ..., I_{t-1}, I_{t+1}, ... \}\right), I_{t}\right)
\label{}
\end{equation}
Similar to the prediction task, it also leverages the temporal relationships in the video, but the difference lies in this task can learn better high-level semantics and temporal context. 

\textbf{Jigsaw puzzles} have recently been applied as a pretext task in semi-supervised VAD\cite{wang2022video, shi2023video, barbalau2023ssmtl++}. The main process involves creating jigsaw puzzles by performing temporal, spatial, or spatio-temporal shuffling, and then designing networks to predict the relative or absolute permutation in time, space, or both. The optimization function is as follows,
\begin{equation}
 l_{jig}= \sum_{i}\mathcal{L}\left(t_i, \hat{t}_i\right)
\label{}
\end{equation}
where $t_i$ and $\hat{t}_i$ denote the ground-truth and predicted positions of the $i-$th data in the original sequence, respectively. Unlike the previous pretext tasks, which involve high-quality image generation, jigsaw puzzles are cast as the multi-label classification, enhancing computational efficiency and learning more contextual details.

\textbf{Contrastive learning} is a key approach in self-supervised learning, where the goal is to learn useful representations by distinguishing between similar and dissimilar pairs.  
For semi-supervised VAD, two samples are regarded as a positive pair if they originate from the same sample, and otherwise as a negative sample pair\cite{huang2021abnormal}. We show the contrastive loss as below,
\begin{equation}
    l_{con} = \sum_{i}-\log \frac{\exp(\mathrm{sim}(x_i, x^{+}_i) / \tau)}{\sum_{k} \exp(\mathrm{sim}(x_i, x^{-}_k) / \tau)}
\end{equation}
$x_i$ and $x^+_i$ are the positive pair, and $x_i$ and $x^-_k$ are the negative pair. $\text{sim}(\cdot, \cdot)$ is the similarity function (e.g., cosine similarity). Wang et al. \cite{wang2020cluster} introduced a cluster attention contrast framework for VAD, which is built on top of contrastive learning. During the inference stage, the highest similarity between the test sample and its variants is regarded as the regularity score. Lu et al.\cite{lu2022learnable} further proposed a learnable locality-sensitive hashing with contrastive learning strategy for VAD. 

\textbf{Denoising}\cite{sun2020adversarial, chen2021nm} is very similar to reconstruction, with the main difference being that noise $\eta$ is added to the input data, and the network is encouraged to achieve a denoising effect for the reconstructed data. The benefit is that it can enhance the robustness of network for VAD. The optimization objective is expressed as, 
\begin{equation}
 l_{den}= \mathcal{L}\left(\Phi\left(\theta, x+\eta\right), x\right)
\label{}
\end{equation}

\textbf{Deep sparse coding} is encouraged by the success of traditional sparse reconstruction based VAD methods\cite{cong2011sparse}, upgrade versions leverage deep neural networks for semi-supervised VAD. Unlike the aforementioned reconstruction or prediction tasks, sparse coding typically uses extracted high-level representations rather than raw video image data as input. By learning from a large amount of normal representations, a dictionary of normal patterns is constructed. The total objective is listed as, 
\begin{equation}
 l_{spa}= \|x-Bz\|^2_2+\|z\|_1
\label{}
\end{equation}
Different normal events can be reconstructed through the dictionary $B$ multiplied by the sparse coefficient $z$. For anomalies, it is hard to reconstruct them using the linear combination of elements from the normal dictionary with a sparse coefficient. To overcome the time-consuming inference and low-level hand-crafted features of traditional sparse reconstruction based methods, deep sparse coding based methods are emerged\cite{luo2017revisit,  zhou2019anomalynet, wu2020fast, luo2019video}, simultaneously leveraging the powerful representation capabilities of DNNs and sparse representation techniques to improve detection performance and efficiency.

\textbf{Patch inpainting} involves the process of reconstructing missing or corrupted regions by inferring the missing parts from the available data. This technique mainly leverages the spatial and temporal context to predict the content of the missing regions, ensuring that the reconstructed regions blend seamlessly with the surrounding regions. The optimization objective for patch inpainting can be defined to minimize the difference between the original and the reconstructed patches,

\begin{equation}
 l_{pat}= \mathcal{L}\left(\Phi\left(\theta, x\odot M\right), x\odot \bar{M}\right)
\label{}
\end{equation}
$M$ denotes a mask, where the value of 0 in the mask indicates that the position needs to be inpainted, while a value of 1 indicates that the position does not need to be inpainted, and $\bar{M}$ is a reversed $M$.  
Different from prediction and visual cloze test, patch inpainting takes into greater account the spatial or spatio-temporal context. Zavrtanik et al.\cite{zavrtanik2021reconstruction} regarded anomaly detection as a reconstruction-by-inpainting task, further randomly removed partial image regions and reconstructed the image from partial inpaintings. Then, Ristea et al.\cite{ristea2022self, madan2023self} presented a novel self-supervised predictive architectural building block, a plug-and-play design that can easily be incorporated into various anomaly detection methods. More recently, a self-distilled masked auto-encoder\cite{ristea2024self} was proposed to inpaint original frames.

\textbf{Multiple task} can ease the dilemma of the single pretext task, i.e., a single task may be not well aligned with the VAD task, thus leading to sub-optimal performance. Recently, several works attempted to train VAD models jointly on multiple pretext tasks. For example, various studies exploited different self-supervised task compositions, involving reconstruction and prediction\cite{zhao2017spatio, morais2019learning,ye2019anopcn, bao2022hierarchical}, prediction and denoising\cite{liu2022appearance, liu2022learninga}, prediction and jigsaw puzzles\cite{huang2022self}, prediction and constrastive learning\cite{lu2022learnable}. A few works\cite{georgescu2021anomaly, barbalau2023ssmtl++, shi2023video, zhang2024multi} strove to develop more sophisticated multiple tasks from different perspectives.

\subsubsection{One-class Learning}
One-class learning primarily focuses on samples from the normal class. Compared to self-supervised learning methods, it does not require the strenuous effort of designing feasible pretext tasks. It is generally divided into three categories: one-class classifier, Gaussian classifier, and adversarial classifier from discriminators of generative adversarial networks (GAN).

\textbf{One-class classifier} basically includes one-class  support vector machine (OC-SVM)\cite{scholkopf2001estimating}, support vector data description (SVDD)\cite{tax2004support}, and other extensions, e.g., basic/generalized one-class discriminative subspace classifier (BODS, GODS)\cite{wang2019gods}.  
Specifically, OC-SVM is modeled as an extension of the SVM objective by learning a max-margin hyperplane that separates the normal from the abnormal in a dataset, which is learned by minimizing the following objective,
\begin{equation}
\min_{w,b,\xi\geq 0}\frac{1}{2}\|{w}\|^2_2 - b+C\!\!\sum\xi_i, ~\mathrm{s.t.}~ w^Tx_i\geq b-\xi_i, \forall x_i\in \mathcal{X}_n
\label{eq:oc-svm}
\end{equation}
where $\xi_i$ is the non-negative slack, $w$ and $b$ denote the hyperplane, and $C$ is the slack penalty. AMDN\cite{xu2015learning} is a typical OC-SVM based VAD method, which obtains low-dimensional representations through the auto-encoder, and then uses OC-SVM to classify all normal representations. 
Another popular variant of one-class classifiers is (Deep) SVDD\cite{ruff2018deep, liznerski2021explainable} that instead of modeling data to belong to an open half-space (as in OC-SVM), assumes the normal samples inhabit a bounded set, and the optimization seeks the centroid $c$ of a hypersphere of minimum radius $R>0$ that contains all normal samples. Mathematically, the objective reads,
\begin{equation}
\min_{c,R,\xi\geq 0} \frac{1}{2} R^2\!+C\sum\xi_i, ~\mathrm{s.t.}~ \|x_i-c\|_2^2\!\leq\!R^2-\xi_i, \forall x_i \in \mathcal{X}_n 
\label{eq:svdd}
\end{equation}
where, as in OC-SVM, the $\xi_i$ models the slack. Based on this, Wu et al.\cite{wu2019deep} proposed a end-to-end deep one-class classifier, i.e., DeepOC, for VAD, avoiding the shortcomings of complicated two-stage training of AMDN.

\textbf{Gaussian classifier} based methods\cite{sabokrou2017deep, sabokrou2018deep, fan2020video} assume that, in practical applications, the data typically follows a Gaussian distribution. By using training samples, it learns the Gaussian distribution (mean $\mu$ and variance $\Sigma$) of the normal pattern. During the testing phase, samples that deviate significantly from the mean are considered anomalies. The abnormal score is presented as,
\begin{equation}
    p({x_i}) = \frac{1}{(2\pi)^{\frac{k}{2}} |{\Sigma}|^{\frac{1}{2}}} \exp\left(-\frac{1}{2} ({x_i} - {\mu})^T {\Sigma}^{-1} ({x_i} - {\mu})\right)
\end{equation}

\textbf{Adversarial classifier} uses adversarial training between the generator $G$ and discriminator $D$ to learn the distribution of normal samples. $G$ is aware of normal data distribution, as the normal samples are accessible to it. Therefore, $D$ explicitly decides whether the output of $G$ follows the normal distribution or not. Therefore, adversarial classifier can be jointly learned by optimizing the following objective,
\begin{equation}
\begin{aligned}
\min_{G} \max_{D} ~ & \Big( \mathbb{E}_{x_i \sim  p_t}[\log({D}(x_i))] \\
& + \mathbb{E}_{\tilde{x_i} \sim  p_t+\mathcal{N}_\sigma}[\log(1-{D}({G}(\tilde{x_i})))] \Big), 
\end{aligned}
\label{eq:gan}
\end{equation}
where $x_i$ is drawn from a normal data distribution $p_t$ and $\tilde{x_i}$ is the sample $x_i$ with added noise, which is sampled from a normal distribution $\mathcal{N}_\sigma$. The final abnormal score of an input sample $x$ is given as $D(G(x))$. For instance, Sabokrou et al.\cite{sabokrou2018adversarially, sabokrou2018avid, sabokrou2020deep} developed a conventional adversarial network that contains two sub-networks, where the discriminator works as the one-class classifier, while the refiner supports it by enhancing the normal samples and distorting the anomalies. To mitigate the instability caused by adversarial training, Zaheer et al.\cite{zaheer2020old,zaheer2022stabilizing} proposed stabilizing adversarial classifiers by transforming the role of discriminator to distinguish good and bad quality reconstructions as well as introducing pseudo anomaly examples.

\subsubsection{Interpretable Learning}

\begin{figure}[t]
  \centering
  \includegraphics[width=0.99\linewidth]{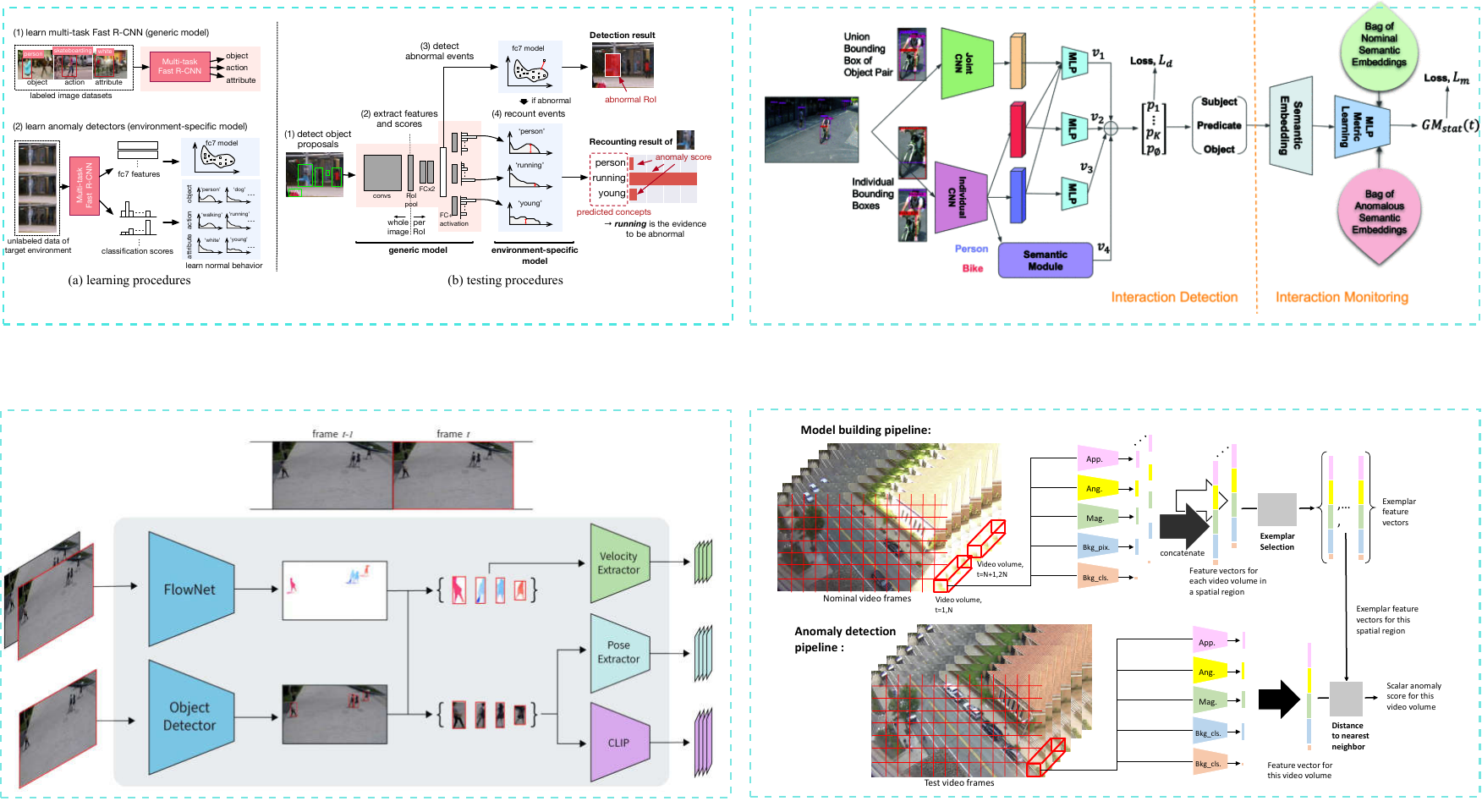}
\caption{Flowchart of four classical explainable VAD methods.}
  \label{fig:interpretable}
\end{figure}

While self-supervised learning and one-class learning based methods perform competitively on popular VAD benchmarks, they are entirely dependent on complex neural networks and mostly trained end-to-end. This limits their interpretability and generalization capacity. Therefore, explainable VAD emerges as a solution, which refers to techniques and methodologies used to identify and explain unusual events in videos. These techniques are designed not only to detect anomalies but also to provide clear explanations for why these anomalies are flagged, which is crucial for trust and transparency in real-world applications. For example, Hinami et al.\cite{hinami2017joint} leveraged multi-task detector as the generic model to learn generic knowledge about visual concepts, e.g., entity, action, and attribute, to describe the events in the human-understandable form, then designed an environment-specific model as the anomaly detector for abnormal event recounting and detection. Similarly, Reiss et al.\cite{reiss2022attribute} extracted explicit attribute-based representations, i.e., velocity and pose, along with implicit semantic representations to make interpretable anomaly decisions. Coincidentally, Doshi and Yilmaz\cite{doshi2023towards} proposed a novel framework which monitors both individuals and the interactions between them, then explores the scene graphs to provide an interpretation for the context of anomalies. Singh et al.\cite{singh2023eval} started a new line for explainable VAD, a more generic model based on high-level appearance and motion features which can provide human-understandable reasons. Compared to previous methods, this work is independent of detectors and is capable of locating spatial anomalies. More recently, Yang et al.\cite{yang2024follow} proposed the first rule-based reasoning framework for semi-supervised VAD with large language models (LLMs) due to LLMs' revolutionary reasoning ability. Here, we present some classical explainable VAD methods in \ref{fig:interpretable}. 

\subsection{Network Architecture}
\subsubsection{Auto-encoder}
It consists of two vital structures, namely, encoder and decoder, in which the encoder compresses the input sample into a latent space representation, significantly reducing the dimensionality of the sample. 
The decoder restores the input sample from the latent space representation, increasing the dimensionality of the sample back to the original input size. Due to its inherent image restoration and high-level representation extraction capabilities, auto-encoder is widely used in image restoration based pretext tasks for self-supervised learning methods, such as reconstruction\cite{hasan2016learning}, prediction\cite{zhao2017spatio}, and inpainting\cite{ristea2024self}. Moreover, auto-encoder is also used to extract features for one-class learning based methods\cite{wu2019deep}, where the extracted features are used for optimizing subsequent one-class classifiers. Auto-encoder structures are highly flexible and can be based on various different base networks, such as 2D CNN\cite{hasan2016learning}, 3D CNN\cite{zhao2017spatio}, recurrent neural network (RNN)\cite{luo2019video}, gated recurrent unit (GRU)\cite{morais2019learning}, long short-term memory (LSTM)\cite{medel2016anomaly, luo2017remembering}, graph convolutional network (GCN)\cite{luo2021normal, li2022human} and Transformer\cite{yang2023video, ristea2024self}.

\subsubsection{GAN} 
GAN has been widely adopted in various applications, including VAD, due to its powerful generative capabilities. The key idea is to use the generator and discriminator to identify abnormal sample that deviate from the learned normal distribution. Specifically, GAN, like auto-encoder, is mainly applied to image restoration based pretext tasks\cite{liu2018future, ravanbakhsh2019training, vu2019robust, song2019learning, feng2021convolutional}, where the generator creates restored images, and the discriminator is discarded once the training is finished. On the contrary, several one-class learning based methods\cite{sabokrou2018adversarially, zaheer2020old} leverage the discriminator to assess the likelihood that a new sample is real (normal) or generated (anomalous). Low likelihood scores indicate anomalies, thereby achieving end-to-end one-class classifiers.

\subsubsection{Diffusion}
Diffusion models have achieved state-of-the-art performance across a wide range of generative tasks and have become a promising research topic. Unlike traditional generative models like GAN or auto-encoder, diffusion models generate samples by reversing a diffusion process that gradually destroys sample structure. This reverse process reconstructs the sample from noise in a step-by-step manner, leading to high-quality results. Therefore, diffusion models also appear in the image restoration based pretext tasks. 
Yan et al.\cite{yan2023feature} and Flaborea et al.\cite{flaborea2023multimodal} introduced novel diffusion-based methods to predict the features using the RGB frame and skeleton respectively for VAD.

\subsection{Model Refinement}
\subsubsection{Pseudo Anomalies}
In semi-supervised learning, there is usually a scarcity of real anomalous samples. To compensate for this lack, several research studies opt to generate pseudo-anomalies. Current approaches include: \textit{Perturbing normal samples}, i.e., applying random perturbations to normal video samples, such as adding noise, shuffling frame sequences, or adding extra patches\cite{wu2023dss, astrid2022limiting, astrid2021learning, astrid2023pseudobound}; \textit{Leveraging generative models}, i.e., using GAN or Diffusion to generate samples similar to normal ones but with anomalous features\cite{zaheer2022stabilizing, pourreza2021g2d}; \textit{Simulating specific anomalous behaviors}, i.e., manually introducing extra anomalous samples at the image-level or feature-level\cite{georgescu2021background, liu2023generating}. As a result, training with pseudo-anomalous samples allows the detection model to learn a variety of anomalous patterns, and helps the model learn a broader range of anomalous features, improving its robustness and generalization abilities in real-world applications.

\subsubsection{Memory Bank}
Memory banks\cite{leng2022anomaly, yu2022effective, liu2023diversity, liu2023stochastic, sun2023learning, wang2023memory} are used to store feature representations of normal video samples, which serve as a reference baseline and can be dynamically updated to adapt to new normal patterns, thereby enabling the model to better capture normal patterns, and simultaneously improving the ability to adapt to changing environments. In specific implementations, memory banks can be combined with different network architectures, such as: auto-encoder based reconstruction (or prediction)\cite{gong2019memorizing, park2020learning, lv2021learning, yang2022dynamic} and 
contrastive learning\cite{cao2024context}.

\begin{figure}[t]
  \centering
  \includegraphics[width=0.99\linewidth]{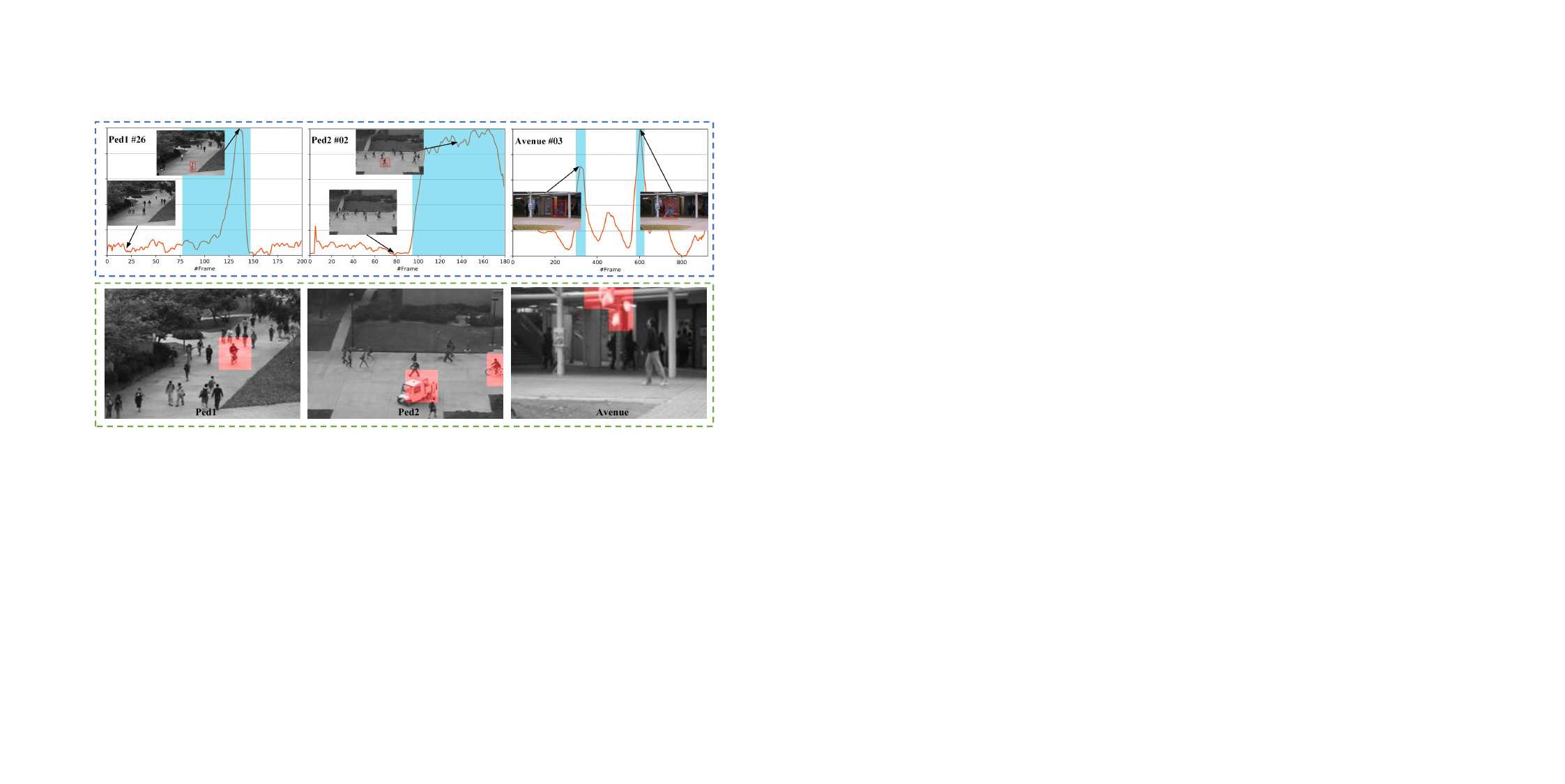}
\caption{Illustrations of frame-level (Top) and pixel-level (Bottom) output.}
  \label{fig: modeloutput}
  \vspace{-0.5cm} 
\end{figure}

\begin{figure*}[t]
  \centering
  \includegraphics[width=0.9\linewidth]{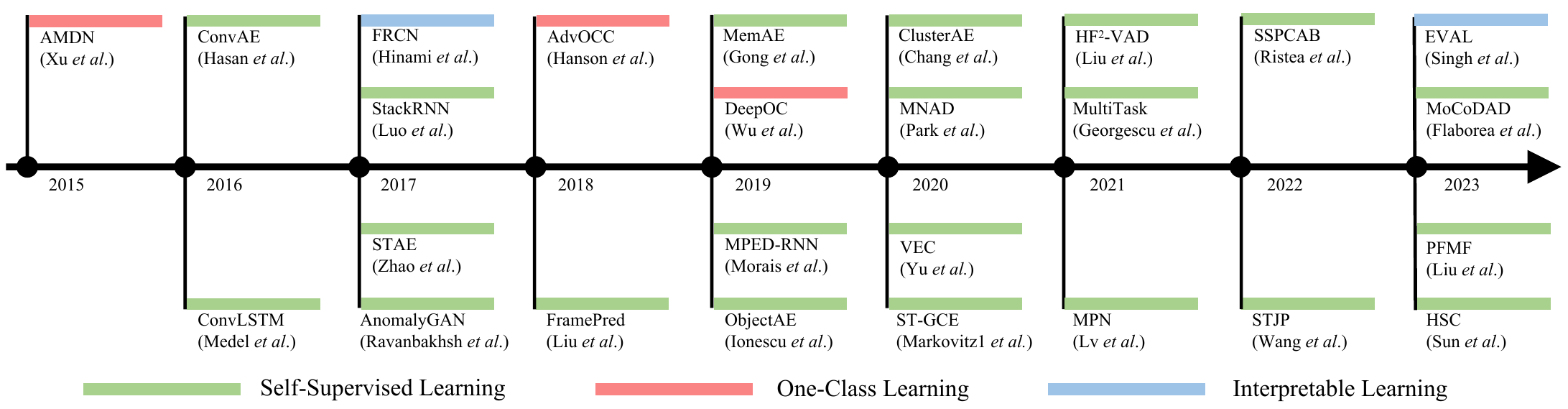}
  \caption{The chronology of representative semi-supervised VAD research.}
  \label{fig:semichronology}
\end{figure*}

\begin{table*}[!t]
    \centering
    
     \caption{Quantitative Performance Comparison of Semi-supervised Methods on Public Datasets.}
    \begin{tabular}{lrl|cccccc}
    \toprule
        \multirow{2}{*}{Method} & \multirow{2}{*}{Publication} & \multirow{2}{*}{Methodology} & Ped1 & Ped2 & Avenue & ShanghaiTech & UBnormal  \\ \cline{4-8}
         &  & & AUC (\%)&AUC (\%)&AUC (\%)&AUC (\%)&AUC (\%) \\ 
        \hline
        AMDN\cite{xu2015learning} & BMVC 2015 & One-class classifier & 92.1 & 90.8 & - & - & - \\ 
        ConvAE\cite{hasan2016learning} & CVPR 2016 & Reconstruction & 81.0 & 90.0 & 72.0 & - & - \\ 
        STAE\cite{zhao2017spatio} & ACMMM 2017 & Hybrid & 92.3 & 91.2 & 80.9 & - & - \\ 
        StackRNN\cite{luo2017revisit} & ICCV 2017 & Sparse coding & - & 92.2 & 81.7 & 68.0 & - \\ 
        FuturePred\cite{liu2018future} & CVPR 2018 & Prediction & 83.1 & 95.4 & 85.1 & 72.8 & - \\ 
        DeepOC\cite{wu2019deep} & TNNLS 2019 & One-class classifier & 83.5 & 96.9 & 86.6 & - & - \\ 
        MemAE\cite{gong2019memorizing} & ICCV 2019 & Reconstruction & - & 94.1 & 83.3 & 71.2 & - \\ 
        AnoPCN\cite{ye2019anopcn} & ACMMM 2019 & Prediction & - & 96.8 & 86.2 & 73.6 & - \\ 
        ObjectAE\cite{ionescu2019object} & CVPR 2019 & One-class classifier & - & 97.8 & 90.4 & 84.9 & - \\
        BMAN\cite{lee2019bman} & TIP 2019 & Prediction & - & 96.6 & 90.0 & 76.2 & - \\ 
        sRNN-AE\cite{luo2019video} & TPAMI 2019 & Sparse coding & - & 92.2 & 83.5 & 69.6 & - \\ 
        ClusterAE\cite{chang2020clustering} & ECCV 2020 & Reconstruction & - & 96.5 & 86.0 & 73.3 & - \\ 
        MNAD\cite{park2020learning} & CVPR 2020 & Reconstruction & ~ & 97.0 & 88.5 & 70.5 & - \\ 
        VEC\cite{yu2020cloze}& ACMMM 2020 & Cloze test & - & 97.3 & 90.2 & 74.8 & - \\ 
        AMMC-Net\cite{cai2021appearance} & AAAI 2021 & Prediction & - & 96.6 & 86.6 & 73.7 & - \\ 
        MPN\cite{lv2021learning} & CVPR 2021 & Prediction & 85.1 & 96.9 & 89.5 & 73.8 & - \\ 
        HF$^2$-VAD\cite{liu2021hybrid} & ICCV 2021 & Hybrid & - & 99.3 & 91.1 & 76.2 & - \\ 
        BAF\cite{georgescu2021background} & TPAMI 2021 & One-class classifier & - & 98.7 & 92.3 & 82.7 & 59.3 \\ 
        Multitask\cite{georgescu2021anomaly} & CVPR 2021 & Multiple tasks & - & 99.8 & 92.8 & 90.2 & - \\ 
        F$^2$PN\cite{luo2021future} & TPAMI 2022 & Prediction & 84.3 & 96.2 & 85.7 & 73.0 & - \\ 
        DLAN-AC\cite{yang2022dynamic} & ECCV 2022 & Reconstruction & - & 97.6 & 89.9 & 74.7 & - \\
        BDPN\cite{chen2022comprehensive} & AAAI 2022 & Prediction & - & 98.3 & 90.3 & 78.1 & - \\ 
        CAFE\cite{yu2022effective} & ACMMM 2022 & Prediction & - & 98.4 & 92.6 & 77.0 & - \\ 
        STJP\cite{wang2022video} & ECCV 2022 & Jigsaw puzzle & - & 99.0 & 92.2 & 84.3 & 56.4 \\ 
        
        MPT\cite{shi2023video} & ICCV 2023 & Multiple tasks & - & 97.6 & 90.9 & 78.8 & - \\
        HSC\cite{sun2023hierarchical} & CVPR 2023 & Hybrid & - & 98.1 & 93.7 & 83.4 & ~ \\ 
        LERF\cite{sun2023learning} & AAAI 2023 & Predicition & - & 99.4 & 91.5 & 78.6 & - \\ 
        DMAD\cite{liu2023diversity} & CVPR 2023 & Reconstruction & - & 99.7 & 92.8 & 78.8 & - \\ 
        EVAL\cite{singh2023eval} & CVPR 2023 & Interpretable learning & - & - & 86.0 & 76.6 & - \\ 
        FBSC-AE\cite{cao2023new} & CVPR 2023 & Prediction & - & - & 86.8 & 79.2 & - \\ 
        FPDM\cite{yan2023feature} & ICCV 2023 & Prediction & - & - & 90.1 & 78.6 & 62.7 \\
        PFMF\cite{liu2023generating} & CVPR 2023 & Multiple tasks & - & - & 93.6 & 85.0 & - \\ 
        STG-NF\cite{hirschorn2023normalizing} & ICCV 2023 & Gaussian classifier & - & - & - & 85.9 & 71.8 \\ 
        AED-MAE\cite{ristea2024self} & CVPR 2024 & Patch inpainting & - & 95.4 & 91.3 & 79.1 & 58.5 \\ 
        SSMCTB\cite{madan2023self} & TPAMI 2024 & Patch inpainting & - & - & 91.6 & 83.7 & - \\ 
        \bottomrule
    
    \end{tabular}
    \label{tab:semiperformance}
    \vspace{-0.5cm}
\end{table*}

\subsection{Model Output}
\subsubsection{Frame-level}
In frame-level output, each frame of the video is classified as normal or abnormal. This output format provides an overall view of which frames in the video contain anomalies. Such output is simple and straightforward, easy to implement and understand, and particularly effective for detecting anomalies over a broad time range.

\subsubsection{Pixel-level}
In pixel-level output, not only is it identified which frames contain anomalies, but also which specific pixel regions within those frames are abnormal\cite{huang2022pixel}. This output format provides more granular information about the anomalies. Pixel-level output offers precise locations and extents of anomalies, providing more detailed information for further analysis of the nature and cause of the anomalies. We illustrate different model output in Figure~\ref{fig: modeloutput}.

\subsection{Performance Comparison}

Figure~\ref{fig:semichronology} presents a concise chronology of semi-supervised VAD methods. Besides, 
Table~\ref{tab:semiperformance} provides a performance summary observed in representative semi-supervised VAD methods.

\section{Weakly Supervised Video Anomaly Detection}\label{weak}
\begin{figure}[t]
  \centering
  \includegraphics[width=0.99\linewidth]{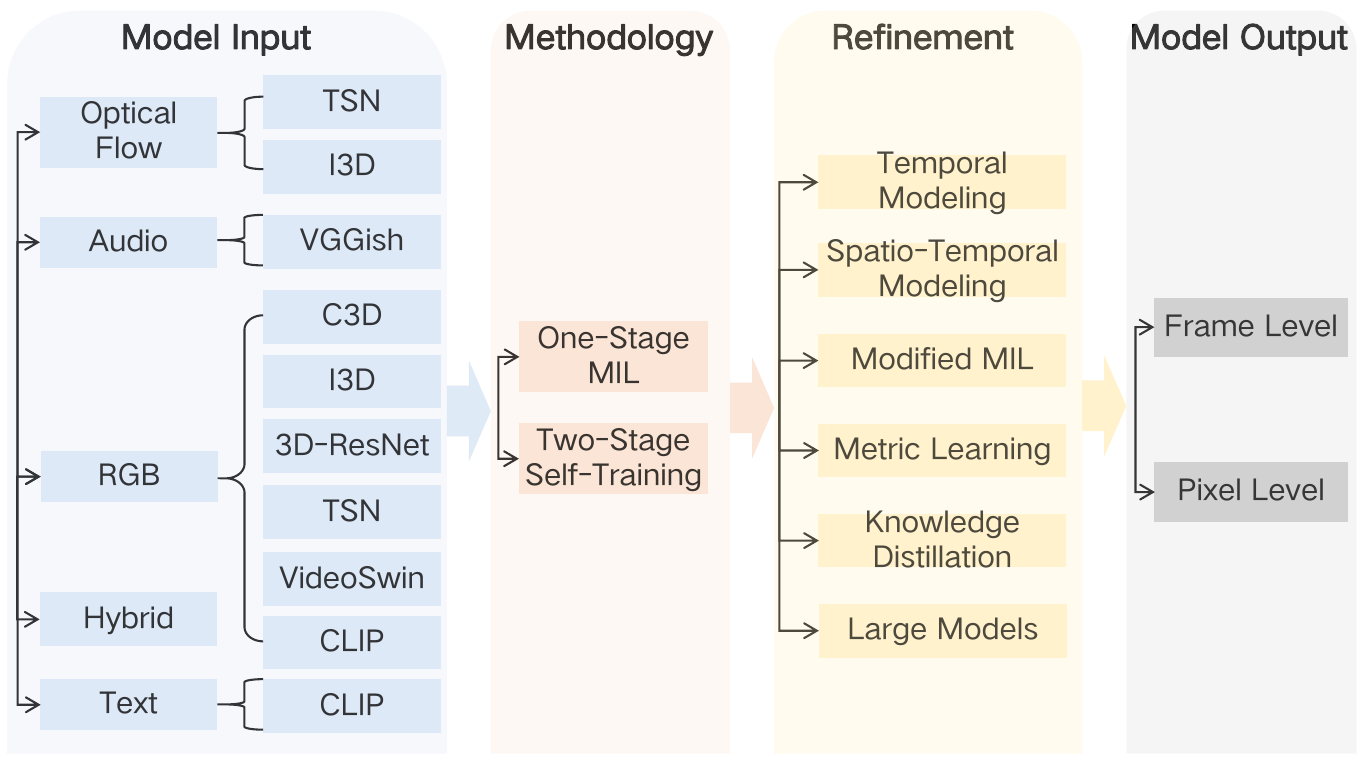}
\caption{The taxonomy of weakly supervised VAD. We provide a hierarchical taxonomy that organizes existing deep weakly supervised VAD models by model input, methodology, refinement strategy and model output into a systematic framework.}
  \label{fig:weaktaxonomy}
  \vspace{-0.6cm} 
\end{figure}

Weakly supervised VAD is currently a highly regarded research direction in the VAD field, with its origins traceable to DeepMIL\cite{sultani2018real}. Compared to semi-supervised VAD, it is a newer research direction, and therefore existing reviews lack a comprehensive and in-depth introduction. As shown in Table~\ref{tab:survey}, both Chandrakala et al.\cite{chandrakala2023anomaly} and Liu et al.\cite{liu2023generalized} mention the weakly supervised VAD task. However, the former only briefly describes several achievements from 2018 to 2020, while the latter, although encompassing recent works, lacks a scientific taxonomy and simply categorizes them into single-modal and multi-modal based on the different modalities. Given this context, we survey related works from 2018 to the present, including the latest methods based on pre-trained large models, and we classify existing works from four aspects: model input, methodology, refinement strategy, and model output. The taxonomy of weakly supervised VAD is illustrated in Figure~\ref{fig:weaktaxonomy}.

Compared to semi-supervised VAD, weakly supervised VAD explicitly defines anomalies during the training process, giving the detection algorithm a clear direction. However, in contrast to fully supervised VAD, the coarse weak supervision signals introduce uncertainty into the detection process. Most existing methods utilize the MIL mechanism to optimize the model. This process can be viewed as selecting the hardest regions (video clips) that appear most abnormal from normal bags (normal videos) and the regions most likely to be abnormal from abnormal bags (abnormal videos). Then, the goal is to maximize the predicted confidence difference between them (with the confidence for the hardest normal regions approaching 0 and the confidence for the most abnormal regions approaching 1), which can be regarded as a binary classification optimization. By gradually mining all normal and abnormal regions based on their different characteristics, the anomaly confidence of abnormal regions increases while that of normal regions decreases. Unfortunately, due to the lack of strong supervision signals, the detection model inevitably involves blind guessing in the above optimization process.

\subsection{Model Input}
Unlike semi-supervised VAD, the network input for weakly supervised VAD is not the raw video, such as RGB, optical flow, or skeleton. Instead, they are features extracted by pre-trained models. This approach alleviates the problem posed by the large scale of existing weakly supervised VAD datasets, the diverse and complex scenes, and the weak supervision signals. Using pre-trained features as input allows the effective utilization of off-the-shelf models' learned knowledge of appearance and motion, significantly reduces the complexity of the detection model, and enables efficient training.

\subsubsection{RGB}
RGB is the most common model input. The general approach divides a long video into multiple segments and uses pre-trained visual models to extract global features from each segment. As deep models continue to evolve and improve, the visual models used have also been upgraded, progressing from the initial C3D\cite{tran2015learning, sultani2018real, zaheer2020claws} to I3D\cite{carreira2017quo, wu2020not, wu2022self, zhou2023batchnorm, almarri2024multi}, 3D-ResNet\cite{hara2018can, sun2023long}, TSN\cite{wang2016temporal, zhong2019graph, li2022weakly}, and more recently, to the popular Swin Transformer\cite{liu2021swin, li2022self} and CLIP\cite{radford2021learning, wu2024open}. This continuous upgrade in visual models has led to a gradual improvement in detection performance. 

\subsubsection{Optical Flow}
Similar to RGB, the same approach is applied to optical flow input to obtain corresponding global features. However, due to the time-consuming nature of optical flow extraction, it is less commonly used in existing methods. Common pre-trained models for optical flow include I3D\cite{wan2020weakly} and TSN\cite{zhong2019graph}.

\subsubsection{Audio}
For multimodal datasets (e.g., XD-Violence) containing audio signals, audio also holds significant perceptual information. Unlike RGB images, audio is one-dimensional and is typically processed as follows, audios are resampled, compute the spectrograms, and create log mel spectrograms, then these features are framed into non-overlapping examples. Finally, these examples are then fed into a pre-trained audio model, such as VGGish\cite{hershey2017cnn}, to extract features\cite{pang2021violence,peng2023learning}. 

\subsubsection{Text}
More recently, several researchers\cite{pu2023learning, chen2023tevad, tao2024learn, wu2024toward} attempt to incorporate text descriptions related to videos to aid in VAD. These texts may be manually annotated or generated by large models. The text data is typically converted into features using the text encoder and then fed into the subsequent detection network.

\subsubsection{Hybrid}
Common hybrid inputs include RGB combined with optical flow\cite{wan2020weakly}, RGB combined with audio\cite{wei2022look, wei2022msaf, yu2022modality}, RGB combined with optical flow and audio\cite{wu2022weakly}, and, more recently, RGB combined with text\cite{yuan2024towards}. 

\subsection{Methodology}
Under the weak supervision, traditional fully supervised methods are no longer adequate. To address this issue, we identify two different approaches: the one-stage MIL and two-stage self-learning.

\subsubsection{One-stage MIL}

The one-stage MIL\cite{sultani2018real, zhu2019motion, zhang2019temporal, liu2022collaborative} is the most commonly used approach for weakly supervised VAD. The basic idea is to first divide long videos into multiple segments, and then use a MIL mechanism to select the most representative samples from these segments. This includes selecting hard examples from normal videos that look most like anomalies and the most likely abnormal examples from the abnormal videos. The model is then optimized by lowering the anomaly confidence of hard examples and increasing the confidence of the most likely abnormal examples. Ultimately, the confidence of model in predicting normal samples gradually decreases, while its confidence in predicting abnormal samples gradually increases, thereby achieving anomaly detection. The advantage of this method lies in its simplicity and ease of implementation. The MIL objective is showcased as,
\begin{equation}
   l_{mil}= \sum_i{\max \left(0,1-\max \Phi\left(\theta, {x}^a_i\right)+\max \Phi\left(\theta, {x}^n_i\right)\right)}
\end{equation}
where ${x}^a$ and ${x}^n$ denote an abnormal video and a normal video respectively.

Additionally, TopK\cite{wu2020not} extends MIL by selecting the top $K$ segments with the highest prediction scores from each video, rather than just the highest-scoring segment, for training. Therefore, MIL can be seen as a special case of TopK. For these TopK segments, compute their average prediction score as the predicted probability \( \hat{y} \),
\begin{equation}
\hat{y} = \sigma\left( \frac{1}{K} \sum_{i\in topk}^{K} \Phi\left(\theta, x_i\right) \right)
\end{equation}
where \( \sigma \) is the sigmoid activation function. 
The cross-entropy loss between $\hat{y}$ and the label $y$ is used to optimize the model,
\begin{equation}
l_{topk} = - \left( y \log(\hat{y}) + (1 - y) \log(1 - \hat{y}) \right)
\end{equation}

The one-stage MIL mechanism leads to models that tend to focus only on the most significant anomalies while ignoring less obvious ones.

\subsubsection{Two-stage Self-training}
In contrast, improved two-stage self-learning is more complex but also more effective. This method employs a two-stage training process. First, a preliminary model is pre-trained using the one-stage MIL. During this phase, the model learns the basic principles of VAD. Then, using the pre-trained model as an initial parameter, a self-learning mechanism is introduced to adaptively train the model further, enhancing its ability to recognize anomalies. Specifically, during the self-learning phase, the predictions of model from the pre-training stage are used to automatically select high-confidence abnormal regions. These regions are then treated as pseudo-labeled data to retrain the model, thereby improving its ability to identify anomalies. This two-stage training approach effectively enhances the performance of model in weakly supervised VAD, further improving the model's generalization ability and robustness. NoiseClearner\cite{zhong2019graph}, MIST\cite{feng2021mist}, MSL\cite{li2022self}, CUPL\cite{zhang2023exploiting}, and TPWNG\cite{yang2024text} are typical two-stage self-training works.

The two-stage self-learning method based on improved MIL excels in weakly supervised VAD, but it also comes with some drawbacks, such as, high computational complexity: the two-stage training process requires more computational resources and time. Both pre-training and self-learning phases involve multiple iterations of training, leading to high computational costs; Dependence on initial model quality: the self-learning stage relies on the initial model generated during pre-training. If the quality of the initial model is poor, erroneous predictions may be treated as pseudo-labels, affecting subsequent training effectiveness. 

\subsection{Refinement Strategy}
Refinement strategies primarily focus on input features, method design, and other aspects to compensate for the shortcomings of weak supervision signals. We compile several commonly used refinement strategies and provide a detailed introduction in this section.

\subsubsection{Temporal Modeling}
Temporal modeling is essential for capturing the critical context information in videos. Unlike actions, anomalous events are complex combinations of scenes, entities, actions, and other elements, which require rich contextual information for accurate reasoning. Existing temporal modeling methods can be broadly categorized into local relationship modeling and global relationship modeling. Local modeling is typically used for online detection\cite{wu2021learning}, whereas global modeling is mainly used for offline detection\cite{tian2021weakly}. Techniques such as temporal convolutional networks\cite{wu2020not, liu2022decouple}, dilated convolution\cite{tian2021weakly}, GCN\cite{chang2021contrastive, cho2023look}, conditional random field\cite{purwanto2021dance}, and transformers\cite{huang2022weakly, li2022self, zhang2022weakly, zhou2023dual} are frequently employed to capture these temporal relationships effectively.

\subsubsection{Spatio-temporal Modeling} Further, spatio-temporal modeling can simultaneously capture spatial relationships, highlighting anomalous spatial locations and effectively reducing noise from irrelevant backgrounds. This can be achieved by segmenting video frames into multiple patches or using existing object detectors to capture foreground objects. Then, methods like self-attention\cite{li2022scale, liu2022learning, sun2023long, ye2024learning} are used to learn the relationships between these patches or objects. Compared to temporal modeling, spatio-temporal modeling involves a higher computational load due to the increased number of entities being analyzed.

\subsubsection{MIL-based Refinement}
The traditional MIL mechanism focuses only on the segments with the highest anomaly scores, which leads to a series of issues, such as ignoring event continuity, the fixed length K-value not adapting to different video scenarios, and bias towards abnormal snippets with simple contexts. Several advanced strategies\cite{lin2019social, park2023normality} aim to address the limitations. By incorporating unbiased MIL\cite{lv2023unbiased}, prior information from text\cite{chen2024prompt, tao2024learn}, magnitude-level MIL\cite{chen2023mgfn}, continuity-aware refinement\cite{gong2022multi}, and adaptive K-values\cite{sapkota2022bayesian}, the detection performance can be significantly improved.

\subsubsection{Feature Metric Learning}
While MIL-based classification ensures the interclass separability of features, this separability at the video level alone is insufficient for accurate anomaly detection. In contrast, enhancing the discriminative power of features through clustering similar features and isolating different ones should complement and even augment the separability achieved by MIL-based classification. Specifically, the basic principle of feature metric learning is to make similar features compact and different features distant in the feature space to improve discrimination. Several works\cite{wu2021learning, huang2022weakly, pu2023learning, zhou2023batchnorm, fioresi2023ted, zaheer2023clustering, tao2024learn} exploited feature metric learning to enhance the feature discrimination.

\subsubsection{Knowledge Distillation}
Knowledge distillation aims to transfer knowledge from the enriched branch to the barren branch to alleviate the semantic gap, which is mainly applied for modality-missing\cite{liu2023distilling} or modality-enhancing\cite{yu2022modality} scenarios.


\begin{table*}[t]
  \centering
  \caption{Quantitative Performance Comparison of Weakly Supervised Methods on Public Datasets.}
  \label{tab:weak}
  \begin{tabular}{@{}lrl|cccc@{}}
  \toprule
  \multirow{2}{*}{Method} & \multirow{2}{*}{Publication}& \multirow{2}{*}{Feature} & {UCF-Crime } & {XD-Violence} & {ShanghaiTech} & {TAD} \\ \cline{4-7}
  & & &AUC (\%)&AP (\%)&AUC (\%)&AUC (\%)\\ 
  \hline
  DeepMIL \cite{sultani2018real}  & CVPR 2018  & C3D$^{RGB}$   & 75.40     & -   & -   & -       \\
  GCN \cite{zhong2019graph} & CVPR 2019 & TSN$^{RGB}$    & 82.12   & -     & 84.44       & -        \\ 
  
  HLNet \cite{wu2020not} &  ECCV 2020  & I3D$^{RGB}$  & 82.44    & 75.41      & -       & -        \\ 
  CLAWS \cite{zaheer2020claws} & ECCV 2020 & C3D$^{RGB}$    & 83.03    & -      & 89.67      & -        \\
  
  MIST \cite{feng2021mist}  &  CVPR 2021& I3D$^{RGB}$   & 82.30    & -    & 94.83       & -\\
  RTFM \cite{tian2021weakly}  &ICCV 2021  & I3D$^{RGB}$   & 84.30    & 77.81     & 97.21       & -        \\
  CTR \cite{wu2021learning} & TIP 2021  & I3D$^{RGB}$   & 84.89    & 75.90     & 97.48       & -        \\ 
  
  MSL \cite{li2022self} & AAAI 2022 & VideoSwin$^{RGB}$    & 85.62   & 78.59     & 97.32       & -    \\ 
  S3R \cite{wu2022self} & ECCV 2022  & I3D$^{RGB}$   & 85.99  & 80.26     & 97.48     & -        \\ 
  SSRL \cite{li2022scale} & ECCV 2022  & I3D$^{RGB}$   & 87.43   & -     & 97.98      & -        \\ 

  CMRL\cite{cho2023look} & CVPR 2023 & I3D$^{RGB}$    & 86.10    & 81.30      & 97.60       & -        \\ 
  CUPL \cite{zhang2023exploiting} & CVPR 2023 & I3D$^{RGB}$    & 86.22    & 81.43      & -       & 91.66        \\ 
  MGFN\cite{chen2023mgfn} &  AAAI 2023 & VideoSwin$^{RGB}$   & 86.67    & 80.11     & -       & -        \\ 
  UMIL \cite{lv2023unbiased} & CVPR 2023 & CLIP    & 86.75   & -      & -      & 92.93     \\ 
  DMU \cite{zhou2023dual} & AAAI 2023 & I3D$^{RGB}$    & 86.97    & 81.66     & -      & -        \\ 
  
  PE-MIL\cite{chen2024prompt} & CVPR 2024 & I3D$^{RGB}$    & 86.83    & 88.05      & 98.35       & -        \\ 
  TPWNG\cite{yang2024text} & CVPR 2024 & CLIP    & 87.79    & 83.68   & -      & -     \\ 
  VadCLIP\cite{wu2024vadclip} &AAAI 2024  & CLIP   & 88.02    & 84.51    & -      & -        \\ 
  STPrompt\cite{wu2024weakly} &ACMMM 2024  & CLIP   & 88.08    & -    & 97.81     & -        \\ 
  \bottomrule
  \end{tabular}
  \end{table*}

\begin{figure*}[t]
  \centering
  \includegraphics[width=0.8\linewidth]{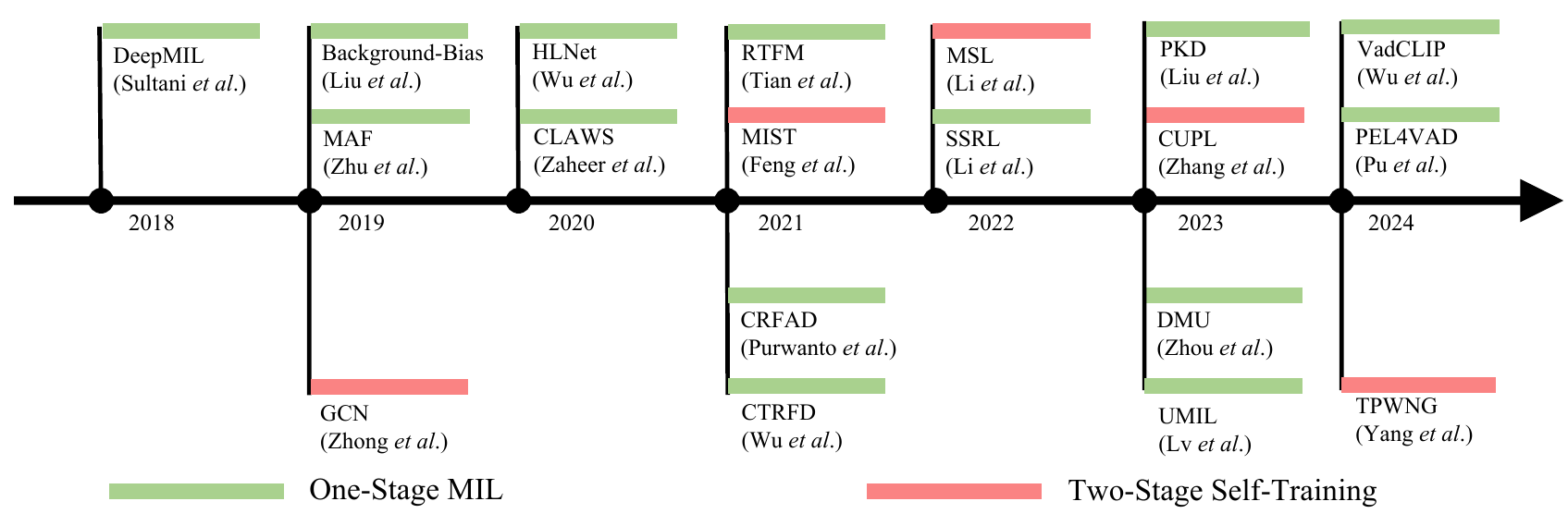}
  \caption{The Chronology for representative weakly supervised VAD research.}
  \label{fig:weakchronology}
\end{figure*}

\subsubsection{{Leveraging Large Models}}
Large models have begun to show tremendous potential and flexibility in the field of VAD. They not only enhance detection capabilities through vision-language features, e.g., CLIP-TSA\cite{joo2023clip} and cross-modal semantic alignment, e.g., VadCLIP\cite{wu2024vadclip}, but also leverage large language models to generate explanatory texts that improve detection accuracy, e.g., TEVAD\cite{chen2023tevad}, UCA\cite{yuan2024towards}, and VAD-Instruct50k\cite{zhang2024holmes}. Furthermore, they can directly use the prior knowledge of large language models for training-free VAD\cite{lv2024video, zanella2024harnessing}, demonstrating advantages in rapid deployment and cost reduction. Furthermore, the superior zero-shot capabilities of these large models may be leveraged for anomaly detection via various other ways, such as AD-oriented prompts \cite{jeong2023winclip,zhou2024anomalyclip} or generic residual learning \cite{zhu2024toward}. These methods collectively advance the development of VAD technology, providing new avenues and tools for achieving more efficient and interpretable VAD.

\subsection{Model Output}

\subsubsection{Frame-level}
Similar to semi-supervised VAD, the output in weakly supervised VAD is typically frame-level prediction results, indicating the probability of each frame being anomalous. This type of output is intuitive and straightforward, which is why it is commonly adopted.
\subsubsection{Pixel-level}
Although frame-level output is intuitive, it lacks interpretability. Therefore, some works have begun to focus on achieving pixel-level detection. For instance, Liu et al.\cite{liu2019exploring} used spatial-level strong supervision signals to achieve spatial localization. Wu et al.\cite{wu2021weakly} took a different approach by not relying on labor-intensive annotations. Instead, they leveraged algorithms such as object detection and tracking, drawing inspiration from video spatio-temporal localization algorithms, and achieved anomaly spatio-temporal localization through spatio-temporal object tube analysis.

\subsection{Performance Comparison}
As depicted in Figure~\ref{fig:weakchronology}, several significant research works have emerged in the field's chronology. Moreover, 
We present an elaborate performance comparison of existing research, as detailed in Table~\ref{tab:weak}.

\section{Fully Supervised Video Anomaly Detection}\label{full}
Fully supervised VAD refers to the task of detecting video anomalies under the condition that the dataset has detailed frame-level or video-level annotations. Here, we consider video violence detection as a fully supervised VAD task. 

\subsection{Approach Categorization}
Video violence detection typically takes the appearance, motion, skeleton, audio, or a combination of these as the input. 
It can be categorized based on the type of input into the following types:

\noindent\textbf{Appearance }input mainly consists of raw RGB images, directly showcasing the visual effect of video frames. This helps the model better understand anomalies that can be directly detected from a visual perspective. 
Many methods\cite{dong2016multi,zhou2017violent, peixoto2019toward, perez2019detection} used RGB features extracted from raw images using pre-trained models as the model input.

\noindent\textbf{Motion }input mainly includes optical flow, optical flow acceleration, and frame differences. These inputs directly showcase the motion state of objects, helping to identify anomalies from the motion perspective that might be difficult to detect visually. Dong et al.\cite{dong2016multi} and Bruno Peixoto et al.\cite{peixoto2020multimodal} used optical flow and optical flow acceleration as input, while Sudhakaran et al.\cite{sudhakaran2017learning} and Hanson et al.\cite{hanson2018bidirectional} employed frame differences as model input.

\noindent\textbf{Skeleton }input can intuitively display the pose state of humans, allowing the model to exclude background interference and focus on human actions. This enables more intuitive and vivid recognition of violent behavior. Su et al.\cite{su2020human} and Singh et al.\cite{singh2018eye} conducted violence detection by studying the interaction relationships between skeletal points.

\noindent\textbf{Audio }input can provide additional information to aid in identifying violent events\cite{peixoto2020multimodal}. This is because certain violent incidents inevitably involve changes in sound, such variations help us better detect violent events, especially when RGB images may not effectively detect due to issues like occlusion.  

\noindent\textbf{Hybrid }input 
combines the strengths of different modalities to better detect violent events. Cheng et al.\cite{cheng2021rwf} utilized RGB images and optical flow as input, while Shang et al.\cite{shang2022multimodal} combined RGB images with audio as input. Garcia et al.\cite{garcia2023human} fed skeleton and frame differences into detection models.

\begin{table*}[t]
    \centering
    \caption{Quantitative Performance Comparison of Fully Supervised Methods on Public Datasets.}
    \label{tab:full}
    \begin{tabular}{lrc|cccc} 
    \toprule
        \multirow{2}{*}{Method} & \multirow{2}{*}{Publication} & \multirow{2}{*}{Model Input} & Hockey Fights & Violent-Flows & RWF-2000 & Crowed Violence \\ \cline{4-7}
        &&&Accuracy (\%)&Accuracy (\%)&Accuracy (\%)&Accuracy (\%)\\
        \hline
        TS-LSTM\cite{dong2016multi} & PR 2016 &RGB+Flow& 93.9  & - & - & - \\ 
        FightNet\cite{zhou2017violent} & JPCS 2017 &RGB+Flow& 97.0 & - & - & -  \\ 
        ConvLSTM\cite{sudhakaran2017learning} & AVSS 2017 &Frame Difference & 97.1& 94.6 & - & -\\
        BiConvLSTM\cite{hanson2018bidirectional} & ECCVW 2018 & Frame Difference &98.1  & 96.3 & - & - \\
        SPIL\cite{su2020human} & ECCV 2020 &Skeleton& 96.8  & - & 89.3 & 94.5 \\ 
        FlowGatedNet\cite{cheng2021rwf} & ICPR 2020 &RGB+Flow& 98.0 &- & 87.3 & 88.9 \\
        X3D\cite{su2022violence} & AVSS 2022 &RGB& - & 98.0 & 94.0 & - \\
        HSCD\cite{garcia2023human} & CVIU 2023 &Skeleton+Frame Difference& 94.5 & - & 90.3 & 94.3 \\ 
        \bottomrule
    \end{tabular}
    \vspace{-0.3cm} 
\end{table*}

\begin{table*}[t]
    \centering
    
    \caption{Quantitative Performance Comparison of Unsupervised Methods on Public Datasets.}
    \begin{tabular}{lrl|ccccccc}
    \toprule
        \multirow{2}{*}{Method} & \multirow{2}{*}{Publication} & \multirow{2}{*}{Methodology} & Avenue & Subway Exit & Ped1 & Ped2 & ShaihaiTech & UMN \\ \cline{4-9}
        &&&AUC (\%)&AUC (\%)&AUC (\%)&AUC (\%)&AUC (\%)&AUC (\%)\\
        
        \hline
        ADF\cite{del2016discriminative} & ECCV 2016 &Change detection & 78.3 & 82.4 &  - & - & - & 91.0  \\ 
        Unmasking\cite{tudor2017unmasking} & ICCV 2017 & Change detection & 80.6 & 86.3 & 68.4 & 82.2 & - & 95.1  \\ 
        MC2ST\cite{liu2018classifier} & BMVC 2018 &Change detection & 84.4 & 93.1 & 71.8 & 87.5 &  -  & -  \\ 
        DAW\cite{wang2018detecting} & ACMMM 2018 &Pseudo label & 85.3 & 84.5 & 77.8 & 96.4 &- & -  \\ 
        STDOR\cite{pang2020self} & CVPR 2020 & Pseudo label & - & 92.7 & 71.7 & 83.2 & - & 97.4  \\
        TMAE\cite{hu2022detecting} & ICME 2022 &Change detection & 89.8 & - & 75.7 & 94.1 & 71.4 & -  \\ 
        CIL\cite{lin2022causal} & AAAI 2022 &Others& 90.3 & 97.6 & 84.9 & 99.4 & - & 100  \\ 
        LBR-SPR\cite{yu2022deep} & CVPR 2022 &Others & 92.8 & - & 81.1 & 97.2 & 72.6 & -  \\ 
        \bottomrule
    \end{tabular}
    \label{tab:uns}
    \vspace{-0.3cm} 
\end{table*}

\subsection{Performance Comparison}
We present the performance comparison of existing fully supervised VAD research in Table~\ref{tab:full}.

\section{Unsupervised Video Anomaly Detection}\label{unsup}
Despite the great popularity of supervised VAD, supervised methods still have shortcomings in practical applications. On the one hand, we cannot clearly define what constitutes normal behavior of real-life human activities in many cases, e.g., running in a sports ground is normal but running in a library is forbidden. On the other hand, it is impractical to know every possible normal event in advance, especially for scientific research. Therefore, VAD in unsupervised environments is of significant research value. 

\subsection{Approach Categorization}
Through an in-depth investigation, we roughly classify the current unsupervised VAD methods into 3 categories: pseudo label, change detection, and others.

\noindent\textbf{Pseudo label based paradigm} is described as follows.  
Wang et al.\cite{wang2018detecting} proposed a two-stage training approach where an auto-encoder is first trained with an adaptive reconstruction loss threshold to estimate normal events from unlabeled videos. These estimated normal events are then used as pseudo-labels to train an OC-SVM, refining the normality model to exclude anomalies and improve detection performance. Pang et al.\cite{pang2020self} introduced a self-training deep ordinal regression method, starting with initial detection using classical one-class algorithms to generate pseudo-labels for anomalous and normal frames. An end-to-end anomaly score learner is then trained iteratively using a self-training strategy that optimizes the detector with newly generated pseudo labels. Zaheer et al.\cite{zaheer2022generative} proposed an unsupervised generative cooperative learning approach, leveraging the low-frequency nature of anomalies for cross-supervision between a generator and a discriminator, each learning from the pseudo-labels of the other. Al-lahham et al.\cite{al2024coarse} presented a coarse-to-fine pseudo-label generation framework using hierarchical divisive clustering for coarse pseudo-labels at the video level and statistical hypothesis testing for fine pseudo-labels at the segment level, training the anomaly detector with the obtained pseudo-labels.

\noindent\textbf{Change detection based paradigm} can be summarized as follows.  
Del Giorno et al.\cite{del2016discriminative} performed change detection in video frames using a simple logistic regression classifier to measure deviations between the data, making comparisons time-independent by randomly ordering frames. Ionescu et al.\cite{tudor2017unmasking} proposed a change detection framework based on the unmasking technique, determining abnormal events by observing changes in classifier accuracy between consecutive events. Liu et al.\cite{liu2018classifier} linked the heuristic unmasking procedure to multiple classifier two-sample tests in statistical machine learning, aiming to improve the unmasking method. Hu et al.\cite{hu2022detecting} introduced a method based on masked auto-encoders\cite{he2022masked}, where the rare and unusual nature of anomalies leads to poor prediction of changing events, allowing for anomaly detection and scoring in unlabeled videos.

\noindent\textbf{Other paradigm} includes the below methods. 
Li et al.\cite{li2021deep} proposed a clustering technique, where an auto-encoder is trained on a subset of normal data and iterates between hypothetical normal candidates based on clustering and representation learning. The reconstruction error is used as a scoring function to assess normality. Lin et al.\cite{lin2022causal} introduced a causal inference framework to reduce the effect of noisy pseudo-labeling, combining long-term temporal context with local image context for anomaly detection. Yu et al.\cite{yu2022deep} highlighted the effectiveness of deep reconstruction for unsupervised VAD, revealing a normality advantage where normal events have lower reconstruction loss. They integrated a novel self-paced refinement scheme into localization-based reconstruction for unsupervised VAD.

\subsection{Performance Comparison}
we present the performance comparison of existing unsupervised VAD research in Table~\ref{tab:uns}.

\section{Open-set Supervised Video Anomaly Detection}\label{open}
\begin{figure*}[t]
  \centering
  \includegraphics[width=0.99\linewidth]{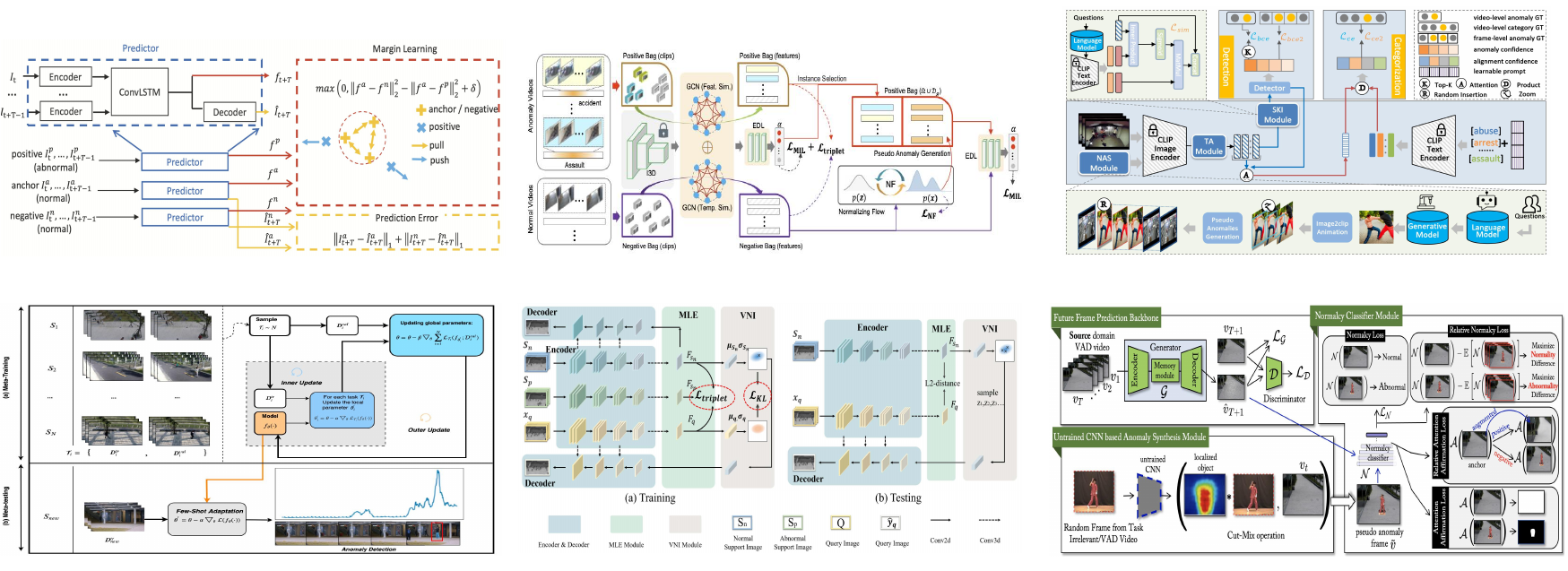}
\caption{Flowchart of six classical open-set supervised VAD methods. Top: open-set methods; Bottom: few-shot methods.}
  \label{fig: open-set}
\end{figure*}

It is challenging to make well-trained supervised model deployed in the open world detect unseen anomalies. Unseen anomalies are highly likely to occur in real-world scenarios, thus research on open-set anomaly detection has garnered significant attention. 
Open-set supervised VAD is a challenging task where the goal is to detect anomalous events in videos that are unseen during the training phase. Unlike traditional (closed-set) VAD, where the types of anomalies are known and well-defined, open-set VAD must handle unforeseen and unknown anomalies. This is crucial for real-world applications, as it is impractical to anticipate and annotate every possible anomaly during training. Therefore, research on open-set VAD has garnered significant attention. 
However, existing review works lack an investigation into open-set VAD. Based on this, we conduct an in-depth survey and make a systematic taxonomy of existing open-set VAD works. To our knowledge, this is the first review that includes a detailed introduction to open-set supervised VAD. In this section, we broadly categorize open-set supervised VAD into two types based on different research directions: open-set VAD and few-set VAD. In Figure~\ref{fig: open-set}, we showcase six classical open-set supervised VAD methods.

\subsection{Open-set VAD}

Open-set VAD is a vital area of research that addresses the limitations of traditional anomaly detection methods by focusing on the detection of unknown anomalies. 
MLEP\cite{liu2019margin}, as the first work on open-set supervised VAD, addressed the challenge of detecting anomalies in videos when only a few anomalous examples are available for training. It emphasizes learning a margin that separates normal and anomalous samples in the feature space. This helps in distinguishing anomalies even when only a few examples are available. 
The follow-up work\cite{acsintoae2022ubnormal} introduced a new benchmark, namely UBnormal, specifically designed for open-set VAD. It aims to provide a comprehensive evaluation framework for testing and comparing various VAD methods under open-set conditions.
Zhu et al.\cite{zhu2022towards} broke through the closed-set detection limitations and developed novel techniques that can generalize to previously unseen anomalies and effectively distinguish them from normal events. Specially, a normalizing flow model is introduced to create pseudo anomaly features. Recently, Wu et al.\cite{wu2024open} extended open-set VAD to the more challenging open-vocabulary VAD, aiming to both detect and recognize anomaly categories. Centered around vision-language models, this task is achieved by matching videos with corresponding textual labels. Additionally, large generative models and language models are utilized to generate pseudo-anomalous samples. There are other methods for the open-set setting, such as \cite{ding2022catching,zhu2024anomaly}, but they are focused on image-level anomaly detection. Through various innovative approaches, such as margin learning, the development of benchmarks, generalization strategies, and the integration of language-vision models, researchers are pushing the boundaries of what is possible in VAD. These advancements are paving the way for more robust, flexible, and practical VAD systems suitable for a wide range of real-world applications. 

\subsection{Few-shot VAD}
The goal of few-shot VAD is to detect anomalies in a previously unseen scene with only a few frames. Compared to open-set VAD, the main difference lies in that a few real frames of unseen anomalies are provided. This task is first introduced by Lu et al.\cite{lu2020few}, and a meta-learning based model is proposed to tackle this problem. During the test stage, the model needs to be fine-tuned via a few provided samples from the new scene. In order to avoid extra fine-tuning processes before deployment, Hu et al.\cite{hu2021adaptive} and Huang et al.\cite{huang2022boosting} adopted the metric-based adaptive network and variational network respectively, both leveraging few normal samples as the reference during the test stage without any fine-tuning. Further, Aich et al.\cite{aich2023cross} presented a novel zxVAD framework, a significant advancement by enabling anomaly detection across domains without requiring target domain adaptation. In this work, a novel untrained CNN based anomaly synthesis module crafts pseudo-abnormal examples by adding foreign objects in normal video frames in a training-free manner. This contrasts with the above few-shot adaptive methods, which require a few labeled data from the target domain for fine-tuning. The former focuses on domain-invariant feature extraction and unsupervised learning, ensuring robustness and generalizability, while the latter relies on few-shot learning to adapt models to new domains with minimal labeled data.

\section{Future Opportunities}
\subsection{Creating Comprehensive Benchmarks}

The current VAD benchmarks have various limitations in terms of data size, modality, and capturing views. Thus, an important future direction is to extend benchmarks along these dimensions for providing more realistic VAD test platforms.
\subsubsection{Large-scale}
Currently, in VAD—especially in semi-supervised VAD—the data scale is too small. For example, the UCSD Ped dataset\cite{li2013anomaly} lasts only a few minutes, and even the larger ShanghaiTech dataset\cite{luo2017remembering} is only a few hours long. Compared to datasets in video action recognition tasks\cite{sun2022human}, which can last hundreds or thousands of hours, VAD datasets are extremely small. This is far from sufficient for training VAD models, as training on small-scale datasets is highly prone to over-fitting in large models. While this might lead to good detection results on the small-scale test data, it can severely impact the performance of VAD models intended for real-world deployment. Therefore, expanding the data scale is a key focus of future research.

\subsubsection{Multi-modal}
Currently, there is limited research on multimodal VAD. 
Just as humans perceive the world through multiple senses\cite{zhu2024vision+}, such as vision, hearing, and smell, effectively utilizing various modality information in the face of multi-source heterogeneous data can enhance the performance of anomaly detection. For example, using audio information can better detect anomalies such as screams and panic, while using infrared information can identify abnormal situations in dark environments.

\subsubsection{Egocentric, Multi-view, 3D, etc.}
Egocentric VAD involves using data captured from wearable devices or body-mounted cameras to simulate how individuals perceive their environment and identify abnormal events, such as detecting falls or aggressive behavior in real time. Creating multi-view benchmarks that leverage data from viewpoints allows for a comprehensive environment analysis, enabling the detection of anomalies that may not be visible from a single perspective. 
3D perspectives from depth information or point clouds can offer more detailed spatial information, enabling models to better understand the structure and context of the environment, which also brings multi-modal signals.

\subsection{Towards Open-world Task}
The current research focuses on the closed-set VAD, which is restricted to detecting only those anomalies that are defined and annotated during training. In applications like urban surveillance, the inability to adapt to unforeseen anomalies limits the practicality and effectiveness of closed-set VAD models. Therefore, moving towards the open-world VAD task, handling the uncertainty and variability of real-world situations, is a feasible future trend. To accomplish this task, several key approaches and their combination can be taken into account.
\textbf{Self-supervised learning: }leveraging unlabeled data to learn discriminative representations that can distinguish between normal and abnormal events\cite{gui2024survey}; \textbf{Open-vocabulary learning: }developing models that can adapt to new anomalies with large models\cite{wu2024open}, pseudo anomaly synthesis, or minimal labeled examples; \textbf{Incremental learning: }continuously updating models with new data and anomaly types without forgetting previously learned information\cite{zhou2024class}.

\subsection{Embracing Pre-trained Large models}
Pre-trained large models have shown remarkable success in various computer vision tasks, and these models can be leveraged in VAD to enhance the understanding and detection of anomalies by integrating semantic context and improving feature representations. Here are several feasible directions. \textbf{Feature extraction:} pre-trained weights of large models, which have been trained on large-scale datasets, provide a strong foundation for feature extraction and reduce the need for extensive training from scratch\cite{joo2023clip}. \textbf{Semantic understanding:} language-vision models can be utilized to understand and incorporate contextual information from video scenes. For instance, text descriptions associated with video frames can provide additional context that helps in identifying anomalies. In the same way, leverage the language capabilities of these models to generate or understand textual descriptions of anomalies, aiding in both the detection and interpretation of the anomalies\cite{zhang2024holmes}. \textbf{Zero-shot learning:} exploit the zero-shot learning capabilities of language-vision models to detect anomalies without requiring explicit examples during training. This is particularly useful in open-set VAD scenarios where new types of anomalies can occur\cite{zhou2024anomalyclip}.

\subsection{Exploiting Interpretable VAD}
Interpretable VAD focuses on creating models that not only detect anomalies but also provide understandable explanations for their predictions. This is crucial for gaining trust in the system, especially in high-stakes applications like surveillance, healthcare, and autonomous vehicles. Here are several feasible directions from three different layers of a VAD system.
\textbf{Input:} instead of directly inputting raw video data into the model, leverage existing technologies to extract key information, such as foreground objects, position coordinates, motion trajectories, and crowd relationships.
\textbf{Algorithm:} combining algorithms from different domains  can be helpful for enhanced reasoning, including:
knowledge graphs, i.e., utilize knowledge graphs to incorporate contextual information and relationships between entities; 
intent prediction, i.e., use intent prediction algorithms to anticipate future actions and detect deviations from expected behaviors\cite{cao2023new}. LLMs' reasoning, i.e., the textual descriptions of detected anomalies using large LLMs can also be used for the explanation. These descriptions can explain what the model perceives as abnormal and why\cite{zhang2024holmes}.
\textbf{Output:} Various aspects such as the spatio-temporal changes and patterns in the video may be synthesized to explain anomalies\cite{wu2024weakly}.

\section{Conclusion}
We present a comprehensive survey of video anomaly detection approaches in the deep learning era. Unlike previous reviews mainly focusing on semi-supervised video anomaly detection, we provide a taxonomy that systematically divides the existing works into five categories by their supervision signals, e.g., semi-supervised, weakly supervised, unsupervised, fully supervised, and open-set supervised video anomaly detection. For each category, we further refine the categories based on model differences, e.g., model input and output, methodology, refinement strategy, and architecture, and we demonstrate the performance comparison of various methods. 
Finally, we discuss several promising research directions for deep learning based video anomaly detection in the future.

\bibliographystyle{IEEEtran}
\bibliography{ref-backup}

\begin{thebibliography}{100}
\providecommand{\url}[1]{#1}
\csname url@samestyle\endcsname
\providecommand{\newblock}{\relax}
\providecommand{\bibinfo}[2]{#2}
\providecommand{\BIBentrySTDinterwordspacing}{\spaceskip=0pt\relax}
\providecommand{\BIBentryALTinterwordstretchfactor}{4}
\providecommand{\BIBentryALTinterwordspacing}{\spaceskip=\fontdimen2\font plus
\BIBentryALTinterwordstretchfactor\fontdimen3\font minus \fontdimen4\font\relax}
\providecommand{\BIBforeignlanguage}[2]{{%
\expandafter\ifx\csname l@#1\endcsname\relax
\typeout{** WARNING: IEEEtran.bst: No hyphenation pattern has been}%
\typeout{** loaded for the language `#1'. Using the pattern for}%
\typeout{** the default language instead.}%
\else
\language=\csname l@#1\endcsname
\fi
#2}}
\providecommand{\BIBdecl}{\relax}
\BIBdecl

\bibitem{pang2021deep}
G.~Pang, C.~Shen, L.~Cao, and A.~V.~D. Hengel, ``Deep learning for anomaly detection: a review,'' \emph{ACM Computing Surveys}, vol.~54, no.~2, pp. 1--38, 2021.

\bibitem{yao2022dota}
Y.~Yao, X.~Wang, M.~Xu, Z.~Pu, Y.~Wang, E.~Atkins, and D.~J. Crandall, ``Dota: unsupervised detection of traffic anomaly in driving videos,'' \emph{IEEE Transactions on Pattern Analysis and Machine Intelligence}, vol.~45, no.~1, pp. 444--459, 2022.

\bibitem{hasan2016learning}
M.~Hasan, J.~Choi, J.~Neumann, A.~K. Roy-Chowdhury, and L.~S. Davis, ``Learning temporal regularity in video sequences,'' in \emph{Proceedings of the IEEE Conference on Computer Vision and Pattern Recognition}, 2016, pp. 733--742.

\bibitem{liu2018future}
W.~Liu, W.~Luo, D.~Lian, and S.~Gao, ``Future frame prediction for anomaly detection--a new baseline,'' in \emph{Proceedings of the IEEE Conference on Computer Vision and Pattern Recognition}, 2018, pp. 6536--6545.

\bibitem{sultani2018real}
W.~Sultani, C.~Chen, and M.~Shah, ``Real-world anomaly detection in surveillance videos,'' in \emph{Proceedings of the IEEE Conference on Computer Vision and Pattern Recognition}, 2018, pp. 6479--6488.

\bibitem{lu2013abnormal}
C.~Lu, J.~Shi, and J.~Jia, ``Abnormal event detection at 150 fps in matlab,'' in \emph{Proceedings of the IEEE International Conference on Computer Vision}, 2013, pp. 2720--2727.

\bibitem{yan2023feature}
C.~Yan, S.~Zhang, Y.~Liu, G.~Pang, and W.~Wang, ``Feature prediction diffusion model for video anomaly detection,'' in \emph{Proceedings of the IEEE International Conference on Computer Vision}, 2023, pp. 5527--5537.

\bibitem{ramachandra2020survey}
B.~Ramachandra, M.~J. Jones, and R.~R. Vatsavai, ``A survey of single-scene video anomaly detection,'' \emph{IEEE Transactions on Pattern Analysis and Machine Intelligence}, vol.~44, no.~5, pp. 2293--2312, 2020.

\bibitem{santhosh2020anomaly}
K.~K. Santhosh, D.~P. Dogra, and P.~P. Roy, ``Anomaly detection in road traffic using visual surveillance: a survey,'' \emph{ACM Computing Surveys}, vol.~53, no.~6, pp. 1--26, 2020.

\bibitem{nayak2021comprehensive}
R.~Nayak, U.~C. Pati, and S.~K. Das, ``A comprehensive review on deep learning-based methods for video anomaly detection,'' \emph{Image and Vision Computing}, vol. 106, p. 104078, 2021.

\bibitem{tran2022anomaly}
T.~M. Tran, T.~N. Vu, N.~D. Vo, T.~V. Nguyen, and K.~Nguyen, ``Anomaly analysis in images and videos: a comprehensive review,'' \emph{ACM Computing Surveys}, vol.~55, no.~7, pp. 1--37, 2022.

\bibitem{chandrakala2023anomaly}
S.~Chandrakala, K.~Deepak, and G.~Revathy, ``Anomaly detection in surveillance videos: a thematic taxonomy of deep models, review and performance analysis,'' \emph{Artificial Intelligence Review}, vol.~56, no.~4, pp. 3319--3368, 2023.

\bibitem{liu2023generalized}
Y.~Liu, D.~Yang, Y.~Wang, J.~Liu, J.~Liu, A.~Boukerche, P.~Sun, and L.~Song, ``Generalized video anomaly event detection: systematic taxonomy and comparison of deep models,'' \emph{ACM Computing Surveys}, vol.~56, no.~7, 2023.

\bibitem{luo2017remembering}
W.~Luo, W.~Liu, and S.~Gao, ``Remembering history with convolutional lstm for anomaly detection,'' in \emph{Proceedings of the IEEE International Conference on Multimedia and Expo}, 2017, pp. 439--444.

\bibitem{zhao2017spatio}
Y.~Zhao, B.~Deng, C.~Shen, Y.~Liu, H.~Lu, and X.-S. Hua, ``Spatio-temporal autoencoder for video anomaly detection,'' in \emph{Proceedings of the ACM International Conference on Multimedia}, 2017, pp. 1933--1941.

\bibitem{ravanbakhsh2017abnormal}
M.~Ravanbakhsh, M.~Nabi, E.~Sangineto, L.~Marcenaro, C.~Regazzoni, and N.~Sebe, ``Abnormal event detection in videos using generative adversarial nets,'' in \emph{Proceedings of the IEEE International Conference on Image Processing}, 2017, pp. 1577--1581.

\bibitem{nguyen2019anomaly}
T.-N. Nguyen and J.~Meunier, ``Anomaly detection in video sequence with appearance-motion correspondence,'' in \emph{Proceedings of the IEEE International Conference on Computer Vision}, 2019, pp. 1273--1283.

\bibitem{xu2015learning}
D.~Xu, E.~Ricci, Y.~Yan, J.~Song, and N.~Sebe, ``Learning deep representations of appearance and motion for anomalous event detection,'' in \emph{Proceedings of the British Machine Vision Conference}, 2015.

\bibitem{xu2017detecting}
D.~Xu, Y.~Yan, E.~Ricci, and N.~Sebe, ``Detecting anomalous events in videos by learning deep representations of appearance and motion,'' \emph{Computer Vision and Image Understanding}, vol. 156, pp. 117--127, 2017.

\bibitem{wu2019deep}
P.~Wu, J.~Liu, and F.~Shen, ``A deep one-class neural network for anomalous event detection in complex scenes,'' \emph{IEEE Transactions on Neural Networks and Learning Systems}, vol.~31, no.~7, pp. 2609--2622, 2019.

\bibitem{sabokrou2017deep}
M.~Sabokrou, M.~Fayyaz, M.~Fathy, and R.~Klette, ``Deep-cascade: cascading 3d deep neural networks for fast anomaly detection and localization in crowded scenes,'' \emph{IEEE Transactions on Image Processing}, vol.~26, no.~4, pp. 1992--2004, 2017.

\bibitem{wang2018generative}
T.~Wang, M.~Qiao, Z.~Lin, C.~Li, H.~Snoussi, Z.~Liu, and C.~Choi, ``Generative neural networks for anomaly detection in crowded scenes,'' \emph{IEEE Transactions on Information Forensics and Security}, vol.~14, no.~5, pp. 1390--1399, 2018.

\bibitem{fan2020video}
Y.~Fan, G.~Wen, D.~Li, S.~Qiu, M.~D. Levine, and F.~Xiao, ``Video anomaly detection and localization via gaussian mixture fully convolutional variational autoencoder,'' \emph{Computer Vision and Image Understanding}, vol. 195, p. 102920, 2020.

\bibitem{hinami2017joint}
R.~Hinami, T.~Mei, and S.~Satoh, ``Joint detection and recounting of abnormal events by learning deep generic knowledge,'' in \emph{Proceedings of the IEEE International Conference on Computer Vision}, 2017, pp. 3619--3627.

\bibitem{ionescu2019object}
R.~T. Ionescu, F.~S. Khan, M.-I. Georgescu, and L.~Shao, ``Object-centric auto-encoders and dummy anomalies for abnormal event detection in video,'' in \emph{Proceedings of the IEEE Conference on Computer Vision and Pattern Recognition}, 2019, pp. 7842--7851.

\bibitem{liu2021hybrid}
Z.~Liu, Y.~Nie, C.~Long, Q.~Zhang, and G.~Li, ``A hybrid video anomaly detection framework via memory-augmented flow reconstruction and flow-guided frame prediction,'' in \emph{Proceedings of the IEEE International Conference on Computer Vision}, 2021, pp. 13\,588--13\,597.

\bibitem{bao2022hierarchical}
Q.~Bao, F.~Liu, Y.~Liu, L.~Jiao, X.~Liu, and L.~Li, ``Hierarchical scene normality-binding modeling for anomaly detection in surveillance videos,'' in \emph{Proceedings of the ACM International Conference on Multimedia}, 2022, pp. 6103--6112.

\bibitem{chen2022comprehensive}
C.~Chen, Y.~Xie, S.~Lin, A.~Yao, G.~Jiang, W.~Zhang, Y.~Qu, R.~Qiao, B.~Ren, and L.~Ma, ``Comprehensive regularization in a bi-directional predictive network for video anomaly detection,'' in \emph{Proceedings of the AAAI Conference on Artificial Intelligence}, vol.~36, no.~1, 2022, pp. 230--238.

\bibitem{sun2022evidential}
C.~Sun, Y.~Jia, and Y.~Wu, ``Evidential reasoning for video anomaly detection,'' in \emph{Proceedings of the ACM International Conference on Multimedia}, 2022, pp. 2106--2114.

\bibitem{sun2023hierarchical}
S.~Sun and X.~Gong, ``Hierarchical semantic contrast for scene-aware video anomaly detection,'' in \emph{Proceedings of the IEEE Conference on Computer Vision and Pattern Recognition}, 2023, pp. 22\,846--22\,856.

\bibitem{luo2021future}
W.~Luo, W.~Liu, D.~Lian, and S.~Gao, ``Future frame prediction network for video anomaly detection,'' \emph{IEEE Transactions on Pattern Analysis and Machine Intelligence}, vol.~44, no.~11, pp. 7505--7520, 2021.

\bibitem{wu2020fast}
P.~Wu, J.~Liu, M.~Li, Y.~Sun, and F.~Shen, ``Fast sparse coding networks for anomaly detection in videos,'' \emph{Pattern Recognition}, vol. 107, p. 107515, 2020.

\bibitem{cai2021appearance}
R.~Cai, H.~Zhang, W.~Liu, S.~Gao, and Z.~Hao, ``Appearance-motion memory consistency network for video anomaly detection,'' in \emph{Proceedings of the AAAI Conference on Artificial Intelligence}, vol.~35, no.~2, 2021, pp. 938--946.

\bibitem{liu2022learning}
Y.~Liu, J.~Liu, X.~Zhu, D.~Wei, X.~Huang, and L.~Song, ``Learning task-specific representation for video anomaly detection with spatial-temporal attention,'' in \emph{Proceedings of the IEEE International Conference on Acoustics, Speech, and Signal Processing}, 2022, pp. 2190--2194.

\bibitem{huang2023video}
X.~Huang, C.~Zhao, and Z.~Wu, ``A video anomaly detection framework based on appearance-motion semantics representation consistency,'' in \emph{Proceedings of the IEEE International Conference on Acoustics, Speech, and Signal Processing}, 2023, pp. 1--5.

\bibitem{li2020spatial}
N.~Li, F.~Chang, and C.~Liu, ``Spatial-temporal cascade autoencoder for video anomaly detection in crowded scenes,'' \emph{IEEE Transactions on Multimedia}, vol.~23, pp. 203--215, 2020.

\bibitem{ramachandra2020learning}
B.~Ramachandra, M.~Jones, and R.~Vatsavai, ``Learning a distance function with a siamese network to localize anomalies in videos,'' in \emph{Proceedings of the IEEE Winter Conference on Applications of Computer Vision}, 2020, pp. 2598--2607.

\bibitem{reiss2022attribute}
T.~Reiss and Y.~Hoshen, ``Attribute-based representations for accurate and interpretable video anomaly detection,'' \emph{arXiv preprint arXiv:2212.00789}, 2022.

\bibitem{morais2019learning}
R.~Morais, V.~Le, T.~Tran, B.~Saha, M.~Mansour, and S.~Venkatesh, ``Learning regularity in skeleton trajectories for anomaly detection in videos,'' in \emph{Proceedings of the IEEE Conference on Computer Vision and Pattern Recognition}, 2019, pp. 11\,996--12\,004.

\bibitem{markovitz2020graph}
A.~Markovitz, G.~Sharir, I.~Friedman, L.~Zelnik-Manor, and S.~Avidan, ``Graph embedded pose clustering for anomaly detection,'' in \emph{Proceedings of the IEEE Conference on Computer Vision and Pattern Recognition}, 2020, pp. 10\,539--10\,547.

\bibitem{rodrigues2020multi}
R.~Rodrigues, N.~Bhargava, R.~Velmurugan, and S.~Chaudhuri, ``Multi-timescale trajectory prediction for abnormal human activity detection,'' in \emph{Proceedings of the IEEE Winter Conference on Applications of Computer Vision}, 2020, pp. 2626--2634.

\bibitem{luo2021normal}
W.~Luo, W.~Liu, and S.~Gao, ``Normal graph: spatial temporal graph convolutional networks based prediction network for skeleton based video anomaly detection,'' \emph{Neurocomputing}, vol. 444, pp. 332--337, 2021.

\bibitem{zeng2021hierarchical}
X.~Zeng, Y.~Jiang, W.~Ding, H.~Li, Y.~Hao, and Z.~Qiu, ``A hierarchical spatio-temporal graph convolutional neural network for anomaly detection in videos,'' \emph{IEEE Transactions on Circuits and Systems for Video Technology}, vol.~33, no.~1, pp. 200--212, 2021.

\bibitem{yang2022two}
Y.~Yang, Z.~Fu, and S.~M. Naqvi, ``A two-stream information fusion approach to abnormal event detection in video,'' in \emph{Proceedings of the IEEE International Conference on Acoustics, Speech, and Signal Processing}, 2022, pp. 5787--5791.

\bibitem{li2022human}
N.~Li, F.~Chang, and C.~Liu, ``Human-related anomalous event detection via spatial-temporal graph convolutional autoencoder with embedded long short-term memory network,'' \emph{Neurocomputing}, vol. 490, pp. 482--494, 2022.

\bibitem{huang2022hierarchical}
C.~Huang, Y.~Liu, Z.~Zhang, C.~Liu, J.~Wen, Y.~Xu, and Y.~Wang, ``Hierarchical graph embedded pose regularity learning via spatio-temporal transformer for abnormal behavior detection,'' in \emph{Proceedings of the ACM International Conference on Multimedia}, 2022, pp. 307--315.

\bibitem{hirschorn2023normalizing}
O.~Hirschorn and S.~Avidan, ``Normalizing flows for human pose anomaly detection,'' in \emph{Proceedings of the IEEE International Conference on Computer Vision}, 2023, pp. 13\,545--13\,554.

\bibitem{yu2023regularity}
S.~Yu, Z.~Zhao, H.~Fang, A.~Deng, H.~Su, D.~Wang, W.~Gan, C.~Lu, and W.~Wu, ``Regularity learning via explicit distribution modeling for skeletal video anomaly detection,'' \emph{IEEE Transactions on Circuits and Systems for Video Technology}, pp. 1--1, 2023.

\bibitem{flaborea2023multimodal}
A.~Flaborea, L.~Collorone, G.~M.~D. Di~Melendugno, S.~D'Arrigo, B.~Prenkaj, and F.~Galasso, ``Multimodal motion conditioned diffusion model for skeleton-based video anomaly detection,'' in \emph{Proceedings of the IEEE International Conference on Computer Vision}, 2023, pp. 10\,318--10\,329.

\bibitem{stergiou2024holistic}
A.~Stergiou, B.~De~Weerdt, and N.~Deligiannis, ``Holistic representation learning for multitask trajectory anomaly detection,'' in \emph{Proceedings of the IEEE Winter Conference on Applications of Computer Vision}, 2024, pp. 6729--6739.

\bibitem{pi2024eogt}
R.~Pi, P.~Wu, X.~He, and Y.~Peng, ``Eogt: video anomaly detection with enhanced object information and global temporal dependency,'' \emph{ACM Transactions on Multimedia Computing, Communications and Applications}, 2024.

\bibitem{chang2020clustering}
Y.~Chang, Z.~Tu, W.~Xie, and J.~Yuan, ``Clustering driven deep autoencoder for video anomaly detection,'' in \emph{Proceedings of the European Conference on Computer Vision}, 2020, pp. 329--345.

\bibitem{fang2020anomaly}
Z.~Fang, J.~Liang, J.~T. Zhou, Y.~Xiao, and F.~Yang, ``Anomaly detection with bidirectional consistency in videos,'' \emph{IEEE Transactions on Neural Networks and Learning Systems}, vol.~33, no.~3, pp. 1079--1092, 2020.

\bibitem{huang2021self}
C.~Huang, Z.~Yang, J.~Wen, Y.~Xu, Q.~Jiang, J.~Yang, and Y.~Wang, ``Self-supervision-augmented deep autoencoder for unsupervised visual anomaly detection,'' \emph{IEEE Transactions on Cybernetics}, vol.~52, no.~12, pp. 13\,834--13\,847, 2021.

\bibitem{zhou2019attention}
J.~T. Zhou, L.~Zhang, Z.~Fang, J.~Du, X.~Peng, and Y.~Xiao, ``Attention-driven loss for anomaly detection in video surveillance,'' \emph{IEEE Transactions on Circuits and Systems for Video Technology}, vol.~30, no.~12, pp. 4639--4647, 2019.

\bibitem{zhang2020normality}
Y.~Zhang, X.~Nie, R.~He, M.~Chen, and Y.~Yin, ``Normality learning in multispace for video anomaly detection,'' \emph{IEEE Transactions on Circuits and Systems for Video Technology}, vol.~31, no.~9, pp. 3694--3706, 2020.

\bibitem{wang2021robust}
X.~Wang, Z.~Che, B.~Jiang, N.~Xiao, K.~Yang, J.~Tang, J.~Ye, J.~Wang, and Q.~Qi, ``Robust unsupervised video anomaly detection by multipath frame prediction,'' \emph{IEEE Transactions on Neural Networks and Learning Systems}, vol.~33, no.~6, pp. 2301--2312, 2021.

\bibitem{yu2021abnormal}
J.~Yu, Y.~Lee, K.~C. Yow, M.~Jeon, and W.~Pedrycz, ``Abnormal event detection and localization via adversarial event prediction,'' \emph{IEEE Transactions on Neural Networks and Learning Systems}, vol.~33, no.~8, pp. 3572--3586, 2021.

\bibitem{zhou2022object}
W.~Zhou, Y.~Li, and C.~Zhao, ``Object-guided and motion-refined attention network for video anomaly detection,'' in \emph{Proceedings of the IEEE International Conference on Multimedia and Expo}, 2022, pp. 1--6.

\bibitem{cheng2023spatial}
K.~Cheng, X.~Zeng, Y.~Liu, M.~Zhao, C.~Pang, and X.~Hu, ``Spatial-temporal graph convolutional network boosted flow-frame prediction for video anomaly detection,'' in \emph{Proceedings of the IEEE International Conference on Acoustics, Speech, and Signal Processing}, 2023, pp. 1--5.

\bibitem{liu2023amp}
Y.~Liu, J.~Liu, K.~Yang, B.~Ju, S.~Liu, Y.~Wang, D.~Yang, P.~Sun, and L.~Song, ``Amp-net: appearance-motion prototype network assisted automatic video anomaly detection system,'' \emph{IEEE Transactions on Industrial Informatics}, vol.~20, no.~2, pp. 2843--2855, 2023.

\bibitem{yu2020cloze}
G.~Yu, S.~Wang, Z.~Cai, E.~Zhu, C.~Xu, J.~Yin, and M.~Kloft, ``Cloze test helps: effective video anomaly detection via learning to complete video events,'' in \emph{Proceedings of the ACM International Conference on Multimedia}, 2020, pp. 583--591.

\bibitem{yang2023video}
Z.~Yang, J.~Liu, Z.~Wu, P.~Wu, and X.~Liu, ``Video event restoration based on keyframes for video anomaly detection,'' in \emph{Proceedings of the IEEE Conference on Computer Vision and Pattern Recognition}, 2023, pp. 14\,592--14\,601.

\bibitem{yu2023video}
G.~Yu, S.~Wang, Z.~Cai, X.~Liu, E.~Zhu, and J.~Yin, ``Video anomaly detection via visual cloze tests,'' \emph{IEEE Transactions on Information Forensics and Security}, vol.~18, pp. 4955--4969, 2023.

\bibitem{wang2022video}
G.~Wang, Y.~Wang, J.~Qin, D.~Zhang, X.~Bao, and D.~Huang, ``Video anomaly detection by solving decoupled spatio-temporal jigsaw puzzles,'' in \emph{Proceedings of the European Conference on Computer Vision}, 2022, pp. 494--511.

\bibitem{shi2023video}
C.~Shi, C.~Sun, Y.~Wu, and Y.~Jia, ``Video anomaly detection via sequentially learning multiple pretext tasks,'' in \emph{Proceedings of the IEEE International Conference on Computer Vision}, 2023, pp. 10\,330--10\,340.

\bibitem{barbalau2023ssmtl++}
A.~Barbalau, R.~T. Ionescu, M.-I. Georgescu, J.~Dueholm, B.~Ramachandra, K.~Nasrollahi, F.~S. Khan, T.~B. Moeslund, and M.~Shah, ``Ssmtl++: revisiting self-supervised multi-task learning for video anomaly detection,'' \emph{Computer Vision and Image Understanding}, vol. 229, p. 103656, 2023.

\bibitem{huang2021abnormal}
C.~Huang, Z.~Wu, J.~Wen, Y.~Xu, Q.~Jiang, and Y.~Wang, ``Abnormal event detection using deep contrastive learning for intelligent video surveillance system,'' \emph{IEEE Transactions on Industrial Informatics}, vol.~18, no.~8, pp. 5171--5179, 2021.

\bibitem{wang2020cluster}
Z.~Wang, Y.~Zou, and Z.~Zhang, ``Cluster attention contrast for video anomaly detection,'' in \emph{Proceedings of the ACM International Conference on Multimedia}, 2020, pp. 2463--2471.

\bibitem{lu2022learnable}
Y.~Lu, C.~Cao, Y.~Zhang, and Y.~Zhang, ``Learnable locality-sensitive hashing for video anomaly detection,'' \emph{IEEE Transactions on Circuits and Systems for Video Technology}, vol.~33, no.~2, pp. 963--976, 2022.

\bibitem{sun2020adversarial}
C.~Sun, Y.~Jia, H.~Song, and Y.~Wu, ``Adversarial 3d convolutional auto-encoder for abnormal event detection in videos,'' \emph{IEEE Transactions on Multimedia}, vol.~23, pp. 3292--3305, 2020.

\bibitem{chen2021nm}
D.~Chen, L.~Yue, X.~Chang, M.~Xu, and T.~Jia, ``Nm-gan: noise-modulated generative adversarial network for video anomaly detection,'' \emph{Pattern Recognition}, vol. 116, p. 107969, 2021.

\bibitem{cong2011sparse}
Y.~Cong, J.~Yuan, and J.~Liu, ``Sparse reconstruction cost for abnormal event detection,'' in \emph{Proceedings of the IEEE Conference on Computer Vision and Pattern Recognition}, 2011, pp. 3449--3456.

\bibitem{luo2017revisit}
W.~Luo, W.~Liu, and S.~Gao, ``A revisit of sparse coding based anomaly detection in stacked rnn framework,'' in \emph{Proceedings of the IEEE International Conference on Computer Vision}, 2017, pp. 341--349.

\bibitem{zhou2019anomalynet}
J.~T. Zhou, J.~Du, H.~Zhu, X.~Peng, Y.~Liu, and R.~S.~M. Goh, ``Anomalynet: an anomaly detection network for video surveillance,'' \emph{IEEE Transactions on Information Forensics and Security}, vol.~14, no.~10, pp. 2537--2550, 2019.

\bibitem{luo2019video}
W.~Luo, W.~Liu, D.~Lian, J.~Tang, L.~Duan, X.~Peng, and S.~Gao, ``Video anomaly detection with sparse coding inspired deep neural networks,'' \emph{IEEE Transactions on Pattern Analysis and Machine Intelligence}, vol.~43, no.~3, pp. 1070--1084, 2019.

\bibitem{zavrtanik2021reconstruction}
V.~Zavrtanik, M.~Kristan, and D.~Sko{\v{c}}aj, ``Reconstruction by inpainting for visual anomaly detection,'' \emph{Pattern Recognition}, vol. 112, p. 107706, 2021.

\bibitem{ristea2022self}
N.-C. Ristea, N.~Madan, R.~T. Ionescu, K.~Nasrollahi, F.~S. Khan, T.~B. Moeslund, and M.~Shah, ``Self-supervised predictive convolutional attentive block for anomaly detection,'' in \emph{Proceedings of the IEEE Conference on Computer Vision and Pattern Recognition}, 2022, pp. 13\,576--13\,586.

\bibitem{madan2023self}
N.~Madan, N.-C. Ristea, R.~T. Ionescu, K.~Nasrollahi, F.~S. Khan, T.~B. Moeslund, and M.~Shah, ``Self-supervised masked convolutional transformer block for anomaly detection,'' \emph{IEEE Transactions on Pattern Analysis and Machine Intelligence}, vol.~46, no.~1, pp. 525--542, 2023.

\bibitem{ristea2024self}
N.-C. Ristea, F.-A. Croitoru, R.~T. Ionescu, M.~Popescu, F.~S. Khan, M.~Shah \emph{et~al.}, ``Self-distilled masked auto-encoders are efficient video anomaly detectors,'' in \emph{Proceedings of the IEEE Conference on Computer Vision and Pattern Recognition}, 2024, pp. 15\,984--15\,995.

\bibitem{ye2019anopcn}
M.~Ye, X.~Peng, W.~Gan, W.~Wu, and Y.~Qiao, ``Anopcn: video anomaly detection via deep predictive coding network,'' in \emph{Proceedings of the ACM International Conference on Multimedia}, 2019, pp. 1805--1813.

\bibitem{liu2022appearance}
Y.~Liu, J.~Liu, J.~Lin, M.~Zhao, and L.~Song, ``Appearance-motion united auto-encoder framework for video anomaly detection,'' \emph{IEEE Transactions on Circuits and Systems II: Express Briefs}, vol.~69, no.~5, pp. 2498--2502, 2022.

\bibitem{liu2022learninga}
Y.~Liu, J.~Liu, M.~Zhao, D.~Yang, X.~Zhu, and L.~Song, ``Learning appearance-motion normality for video anomaly detection,'' in \emph{Proceedings of the IEEE International Conference on Multimedia and Expo}, 2022, pp. 1--6.

\bibitem{huang2022self}
C.~Huang, J.~Wen, Y.~Xu, Q.~Jiang, J.~Yang, Y.~Wang, and D.~Zhang, ``Self-supervised attentive generative adversarial networks for video anomaly detection,'' \emph{IEEE Transactions on Neural Networks and Learning Systems}, vol.~34, no.~11, pp. 9389--9403, 2022.

\bibitem{georgescu2021anomaly}
M.-I. Georgescu, A.~Barbalau, R.~T. Ionescu, F.~S. Khan, M.~Popescu, and M.~Shah, ``Anomaly detection in video via self-supervised and multi-task learning,'' in \emph{Proceedings of the IEEE Conference on Computer Vision and Pattern Recognition}, 2021, pp. 12\,742--12\,752.

\bibitem{zhang2024multi}
M.~Zhang, J.~Wang, Q.~Qi, H.~Sun, Z.~Zhuang, P.~Ren, R.~Ma, and J.~Liao, ``Multi-scale video anomaly detection by multi-grained spatio-temporal representation learning,'' in \emph{Proceedings of the IEEE Conference on Computer Vision and Pattern Recognition}, 2024, pp. 17\,385--17\,394.

\bibitem{scholkopf2001estimating}
B.~Sch{\"o}lkopf, J.~C. Platt, J.~Shawe-Taylor, A.~J. Smola, and R.~C. Williamson, ``Estimating the support of a high-dimensional distribution,'' \emph{Neural Computation}, vol.~13, no.~7, pp. 1443--1471, 2001.

\bibitem{tax2004support}
D.~M. Tax and R.~P. Duin, ``Support vector data description,'' \emph{Machine Learning}, vol.~54, pp. 45--66, 2004.

\bibitem{wang2019gods}
J.~Wang and A.~Cherian, ``Gods: generalized one-class discriminative subspaces for anomaly detection,'' in \emph{Proceedings of the IEEE International Conference on Computer Vision}, 2019, pp. 8201--8211.

\bibitem{ruff2018deep}
L.~Ruff, R.~Vandermeulen, N.~Goernitz, L.~Deecke, S.~A. Siddiqui, A.~Binder, E.~M{\"u}ller, and M.~Kloft, ``Deep one-class classification,'' in \emph{Proceedings of the International Conference on Machine Learning}, 2018, pp. 4393--4402.

\bibitem{liznerski2021explainable}
P.~Liznerski, L.~Ruff, R.~A. Vandermeulen, B.~J. Franks, M.~Kloft, and K.-R. M{\"u}ller, ``Explainable deep one-class classification,'' in \emph{Proceedings of the International Conference on Learning Representations}, 2021.

\bibitem{sabokrou2018deep}
M.~Sabokrou, M.~Fayyaz, M.~Fathy, Z.~Moayed, and R.~Klette, ``Deep-anomaly: fully convolutional neural network for fast anomaly detection in crowded scenes,'' \emph{Computer Vision and Image Understanding}, vol. 172, pp. 88--97, 2018.

\bibitem{sabokrou2018adversarially}
M.~Sabokrou, M.~Khalooei, M.~Fathy, and E.~Adeli, ``Adversarially learned one-class classifier for novelty detection,'' in \emph{Proceedings of the IEEE Conference on Computer Vision and Pattern Recognition}, 2018, pp. 3379--3388.

\bibitem{sabokrou2018avid}
M.~Sabokrou, M.~Pourreza, M.~Fayyaz, R.~Entezari, M.~Fathy, J.~Gall, and E.~Adeli, ``Avid: adversarial visual irregularity detection,'' in \emph{Proceedings of the Asian Conference on Computer Vision}, 2018, pp. 488--505.

\bibitem{sabokrou2020deep}
M.~Sabokrou, M.~Fathy, G.~Zhao, and E.~Adeli, ``Deep end-to-end one-class classifier,'' \emph{IEEE Transactions on Neural Networks and Learning Systems}, vol.~32, no.~2, pp. 675--684, 2020.

\bibitem{zaheer2020old}
M.~Z. Zaheer, J.-h. Lee, M.~Astrid, and S.-I. Lee, ``Old is gold: redefining the adversarially learned one-class classifier training paradigm,'' in \emph{Proceedings of the IEEE Conference on Computer Vision and Pattern Recognition}, 2020, pp. 14\,183--14\,193.

\bibitem{zaheer2022stabilizing}
M.~Z. Zaheer, J.-H. Lee, A.~Mahmood, M.~Astrid, and S.-I. Lee, ``Stabilizing adversarially learned one-class novelty detection using pseudo anomalies,'' \emph{IEEE Transactions on Image Processing}, vol.~31, pp. 5963--5975, 2022.

\bibitem{doshi2023towards}
K.~Doshi and Y.~Yilmaz, ``Towards interpretable video anomaly detection,'' in \emph{Proceedings of the IEEE Winter Conference on Applications of Computer Vision}, 2023, pp. 2655--2664.

\bibitem{singh2023eval}
A.~Singh, M.~J. Jones, and E.~G. Learned-Miller, ``Eval: explainable video anomaly localization,'' in \emph{Proceedings of the IEEE Conference on Computer Vision and Pattern Recognition}, 2023, pp. 18\,717--18\,726.

\bibitem{yang2024follow}
Y.~Yang, K.~Lee, B.~Dariush, Y.~Cao, and S.-Y. Lo, ``Follow the rules: reasoning for video anomaly detection with large language models,'' in \emph{Proceedings of the European Conference on Computer Vision}, 2024.

\bibitem{medel2016anomaly}
J.~R. Medel and A.~Savakis, ``Anomaly detection in video using predictive convolutional long short-term memory networks,'' \emph{arXiv preprint arXiv:1612.00390}, 2016.

\bibitem{ravanbakhsh2019training}
M.~Ravanbakhsh, E.~Sangineto, M.~Nabi, and N.~Sebe, ``Training adversarial discriminators for cross-channel abnormal event detection in crowds,'' in \emph{Proceedings of the IEEE Winter Conference on Applications of Computer Vision}, 2019, pp. 1896--1904.

\bibitem{vu2019robust}
H.~Vu, T.~D. Nguyen, T.~Le, W.~Luo, and D.~Phung, ``Robust anomaly detection in videos using multilevel representations,'' in \emph{Proceedings of the AAAI Conference on Artificial Intelligence}, vol.~33, no.~01, 2019, pp. 5216--5223.

\bibitem{song2019learning}
H.~Song, C.~Sun, X.~Wu, M.~Chen, and Y.~Jia, ``Learning normal patterns via adversarial attention-based autoencoder for abnormal event detection in videos,'' \emph{IEEE Transactions on Multimedia}, vol.~22, no.~8, pp. 2138--2148, 2019.

\bibitem{feng2021convolutional}
X.~Feng, D.~Song, Y.~Chen, Z.~Chen, J.~Ni, and H.~Chen, ``Convolutional transformer based dual discriminator generative adversarial networks for video anomaly detection,'' in \emph{Proceedings of the ACM International Conference on Multimedia}, 2021, pp. 5546--5554.

\bibitem{wu2023dss}
P.~Wu, W.~Wang, F.~Chang, C.~Liu, and B.~Wang, ``Dss-net: dynamic self-supervised network for video anomaly detection,'' \emph{IEEE Transactions on Multimedia}, vol.~26, pp. 2124--2136, 2023.

\bibitem{astrid2022limiting}
M.~Astrid, M.~Z. Zaheer, and S.-I. Lee, ``Limiting reconstruction capability of autoencoders using moving backward pseudo anomalies,'' in \emph{Proceedings of the International Conference on Ubiquitous Robots}, 2022, pp. 248--251.

\bibitem{astrid2021learning}
M.~Astrid, M.~Zaheer, J.-Y. Lee, and S.-I. Lee, ``Learning not to reconstruct anomalies,'' in \emph{Proceedings of the British Machine Vision Conference}, 2021.

\bibitem{astrid2023pseudobound}
M.~Astrid, M.~Z. Zaheer, and S.-I. Lee, ``Pseudobound: limiting the anomaly reconstruction capability of one-class classifiers using pseudo anomalies,'' \emph{Neurocomputing}, vol. 534, pp. 147--160, 2023.

\bibitem{pourreza2021g2d}
M.~Pourreza, B.~Mohammadi, M.~Khaki, S.~Bouindour, H.~Snoussi, and M.~Sabokrou, ``G2d: generate to detect anomaly,'' in \emph{Proceedings of the IEEE Winter Conference on Applications of Computer Vision}, 2021, pp. 2003--2012.

\bibitem{georgescu2021background}
M.~I. Georgescu, R.~T. Ionescu, F.~S. Khan, M.~Popescu, and M.~Shah, ``A background-agnostic framework with adversarial training for abnormal event detection in video,'' \emph{IEEE Transactions on Pattern Analysis and Machine Intelligence}, vol.~44, no.~9, pp. 4505--4523, 2021.

\bibitem{liu2023generating}
Z.~Liu, X.-M. Wu, D.~Zheng, K.-Y. Lin, and W.-S. Zheng, ``Generating anomalies for video anomaly detection with prompt-based feature mapping,'' in \emph{Proceedings of the IEEE Conference on Computer Vision and Pattern Recognition}, 2023, pp. 24\,500--24\,510.

\bibitem{leng2022anomaly}
J.~Leng, M.~Tan, X.~Gao, W.~Lu, and Z.~Xu, ``Anomaly warning: learning and memorizing future semantic patterns for unsupervised ex-ante potential anomaly prediction,'' in \emph{Proceedings of the ACM International Conference on Multimedia}, 2022, pp. 6746--6754.

\bibitem{yu2022effective}
G.~Yu, S.~Wang, Z.~Cai, X.~Liu, and C.~Wu, ``Effective video abnormal event detection by learning a consistency-aware high-level feature extractor,'' in \emph{Proceedings of the ACM International Conference on Multimedia}, 2022, pp. 6337--6346.

\bibitem{liu2023diversity}
W.~Liu, H.~Chang, B.~Ma, S.~Shan, and X.~Chen, ``Diversity-measurable anomaly detection,'' in \emph{Proceedings of the IEEE Conference on Computer Vision and Pattern Recognition}, 2023, pp. 12\,147--12\,156.

\bibitem{liu2023stochastic}
Y.~Liu, D.~Yang, G.~Fang, Y.~Wang, D.~Wei, M.~Zhao, K.~Cheng, J.~Liu, and L.~Song, ``Stochastic video normality network for abnormal event detection in surveillance videos,'' \emph{Knowledge-Based Systems}, vol. 280, p. 110986, 2023.

\bibitem{sun2023learning}
C.~Sun, C.~Shi, Y.~Jia, and Y.~Wu, ``Learning event-relevant factors for video anomaly detection,'' in \emph{Proceedings of the AAAI Conference on Artificial Intelligence}, vol.~37, no.~2, 2023, pp. 2384--2392.

\bibitem{wang2023memory}
L.~Wang, J.~Tian, S.~Zhou, H.~Shi, and G.~Hua, ``Memory-augmented appearance-motion network for video anomaly detection,'' \emph{Pattern Recognition}, vol. 138, p. 109335, 2023.

\bibitem{gong2019memorizing}
D.~Gong, L.~Liu, V.~Le, B.~Saha, M.~R. Mansour, S.~Venkatesh, and A.~v.~d. Hengel, ``Memorizing normality to detect anomaly: memory-augmented deep autoencoder for unsupervised anomaly detection,'' in \emph{Proceedings of the IEEE International Conference on Computer Vision}, 2019, pp. 1705--1714.

\bibitem{park2020learning}
H.~Park, J.~Noh, and B.~Ham, ``Learning memory-guided normality for anomaly detection,'' in \emph{Proceedings of the IEEE Conference on Computer Vision and Pattern Recognition}, 2020, pp. 14\,372--14\,381.

\bibitem{lv2021learning}
H.~Lv, C.~Chen, Z.~Cui, C.~Xu, Y.~Li, and J.~Yang, ``Learning normal dynamics in videos with meta prototype network,'' in \emph{Proceedings of the IEEE Conference on Computer Vision and Pattern Recognition}, 2021, pp. 15\,425--15\,434.

\bibitem{yang2022dynamic}
Z.~Yang, P.~Wu, J.~Liu, and X.~Liu, ``Dynamic local aggregation network with adaptive clusterer for anomaly detection,'' in \emph{Proceedings of the European Conference on Computer Vision}, 2022, pp. 404--421.

\bibitem{cao2024context}
C.~Cao, Y.~Lu, and Y.~Zhang, ``Context recovery and knowledge retrieval: a novel two-stream framework for video anomaly detection,'' \emph{IEEE Transactions on Image Processing}, vol.~33, pp. 1810--1825, 2024.

\bibitem{lee2019bman}
S.~Lee, H.~G. Kim, and Y.~M. Ro, ``Bman: bidirectional multi-scale aggregation networks for abnormal event detection,'' \emph{IEEE Transactions on Image Processing}, vol.~29, pp. 2395--2408, 2019.

\bibitem{cao2023new}
C.~Cao, Y.~Lu, P.~Wang, and Y.~Zhang, ``A new comprehensive benchmark for semi-supervised video anomaly detection and anticipation,'' in \emph{Proceedings of the IEEE Conference on Computer Vision and Pattern Recognition}, 2023, pp. 20\,392--20\,401.

\bibitem{huang2022pixel}
C.~Huang, C.~Liu, Z.~Zhang, Z.~Wu, J.~Wen, Q.~Jiang, and Y.~Xu, ``Pixel-level anomaly detection via uncertainty-aware prototypical transformer,'' in \emph{Proceedings of the ACM International Conference on Multimedia}, 2022, pp. 521--530.

\bibitem{tran2015learning}
D.~Tran, L.~Bourdev, R.~Fergus, L.~Torresani, and M.~Paluri, ``Learning spatiotemporal features with 3d convolutional networks,'' in \emph{Proceedings of the IEEE International Conference on Computer Vision}, 2015, pp. 4489--4497.

\bibitem{zaheer2020claws}
M.~Z. Zaheer, A.~Mahmood, M.~Astrid, and S.-I. Lee, ``Claws: clustering assisted weakly supervised learning with normalcy suppression for anomalous event detection,'' in \emph{Proceedings of the European Conference on Computer Vision}, 2020, pp. 358--376.

\bibitem{carreira2017quo}
J.~Carreira and A.~Zisserman, ``Quo vadis, action recognition? a new model and the kinetics dataset,'' in \emph{Proceedings of the IEEE Conference on Computer Vision and Pattern Recognition}, 2017, pp. 6299--6308.

\bibitem{wu2020not}
P.~Wu, J.~Liu, Y.~Shi, Y.~Sun, F.~Shao, Z.~Wu, and Z.~Yang, ``Not only look, but also listen: learning multimodal violence detection under weak supervision,'' in \emph{Proceedings of the European Conference on Computer Vision}, 2020, pp. 322--339.

\bibitem{wu2022self}
J.-C. Wu, H.-Y. Hsieh, D.-J. Chen, C.-S. Fuh, and T.-L. Liu, ``Self-supervised sparse representation for video anomaly detection,'' in \emph{Proceedings of the European Conference on Computer Vision}, 2022, pp. 729--745.

\bibitem{zhou2023batchnorm}
Y.~Zhou, Y.~Qu, X.~Xu, F.~Shen, J.~Song, and H.~Shen, ``Batchnorm-based weakly supervised video anomaly detection,'' \emph{arXiv preprint arXiv:2311.15367}, 2023.

\bibitem{almarri2024multi}
S.~AlMarri, M.~Z. Zaheer, and K.~Nandakumar, ``A multi-head approach with shuffled segments for weakly-supervised video anomaly detection,'' in \emph{Proceedings of the IEEE Winter Conference on Applications of Computer Vision}, 2024, pp. 132--142.

\bibitem{hara2018can}
K.~Hara, H.~Kataoka, and Y.~Satoh, ``Can spatiotemporal 3d cnns retrace the history of 2d cnns and imagenet?'' in \emph{Proceedings of the IEEE Conference on Computer Vision and Pattern Recognition}, 2018, pp. 6546--6555.

\bibitem{sun2023long}
S.~Sun and X.~Gong, ``Long-short temporal co-teaching for weakly supervised video anomaly detection,'' in \emph{Proceedings of the IEEE International Conference on Multimedia and Expo}, 2023, pp. 2711--2716.

\bibitem{wang2016temporal}
L.~Wang, Y.~Xiong, Z.~Wang, Y.~Qiao, D.~Lin, X.~Tang, and L.~Van~Gool, ``Temporal segment networks: towards good practices for deep action recognition,'' in \emph{Proceedings of the European Conference on Computer Vision}, 2016, pp. 20--36.

\bibitem{zhong2019graph}
J.-X. Zhong, N.~Li, W.~Kong, S.~Liu, T.~H. Li, and G.~Li, ``Graph convolutional label noise cleaner: train a plug-and-play action classifier for anomaly detection,'' in \emph{Proceedings of the IEEE Conference on Computer Vision and Pattern Recognition}, 2019, pp. 1237--1246.

\bibitem{li2022weakly}
N.~Li, J.-X. Zhong, X.~Shu, and H.~Guo, ``Weakly-supervised anomaly detection in video surveillance via graph convolutional label noise cleaning,'' \emph{Neurocomputing}, vol. 481, pp. 154--167, 2022.

\bibitem{liu2021swin}
Z.~Liu, Y.~Lin, Y.~Cao, H.~Hu, Y.~Wei, Z.~Zhang, S.~Lin, and B.~Guo, ``Swin transformer: hierarchical vision transformer using shifted windows,'' in \emph{Proceedings of the IEEE International Conference on Computer Vision}, 2021, pp. 10\,012--10\,022.

\bibitem{li2022self}
S.~Li, F.~Liu, and L.~Jiao, ``Self-training multi-sequence learning with transformer for weakly supervised video anomaly detection,'' in \emph{Proceedings of the AAAI Conference on Artificial Intelligence}, vol.~36, no.~2, 2022, pp. 1395--1403.

\bibitem{radford2021learning}
A.~Radford, J.~W. Kim, C.~Hallacy, A.~Ramesh, G.~Goh, S.~Agarwal, G.~Sastry, A.~Askell, P.~Mishkin, J.~Clark \emph{et~al.}, ``Learning transferable visual models from natural language supervision,'' in \emph{Proceedings of the International Conference on Machine Learning}, 2021, pp. 8748--8763.

\bibitem{wu2024open}
P.~Wu, X.~Zhou, G.~Pang, Y.~Sun, J.~Liu, P.~Wang, and Y.~Zhang, ``Open-vocabulary video anomaly detection,'' in \emph{Proceedings of the IEEE Conference on Computer Vision and Pattern Recognition}, 2024, pp. 18\,297--18\,307.

\bibitem{wan2020weakly}
B.~Wan, Y.~Fang, X.~Xia, and J.~Mei, ``Weakly supervised video anomaly detection via center-guided discriminative learning,'' in \emph{Proceedings of the IEEE International Conference on Multimedia and Expo}, 2020, pp. 1--6.

\bibitem{hershey2017cnn}
S.~Hershey, S.~Chaudhuri, D.~P. Ellis, J.~F. Gemmeke, A.~Jansen, R.~C. Moore, M.~Plakal, D.~Platt, R.~A. Saurous, B.~Seybold \emph{et~al.}, ``Cnn architectures for large-scale audio classification,'' in \emph{Proceedings of the IEEE International Conference on Acoustics, Speech, and Signal Processing}, 2017, pp. 131--135.

\bibitem{pang2021violence}
W.-F. Pang, Q.-H. He, Y.-j. Hu, and Y.-X. Li, ``Violence detection in videos based on fusing visual and audio information,'' in \emph{Proceedings of the IEEE International Conference on Acoustics, Speech, and Signal Processing}, 2021, pp. 2260--2264.

\bibitem{peng2023learning}
X.~Peng, H.~Wen, Y.~Luo, X.~Zhou, K.~Yu, Y.~Wang, and Z.~Wu, ``Learning weakly supervised audio-visual violence detection in hyperbolic space,'' \emph{arXiv preprint arXiv:2305.18797}, 2023.

\bibitem{pu2023learning}
Y.~Pu, X.~Wu, and S.~Wang, ``Learning prompt-enhanced context features for weakly-supervised video anomaly detection,'' \emph{arXiv preprint arXiv:2306.14451}, 2023.

\bibitem{chen2023tevad}
W.~Chen, K.~T. Ma, Z.~J. Yew, M.~Hur, and D.~A.-A. Khoo, ``Tevad: improved video anomaly detection with captions,'' in \emph{Proceedings of the IEEE Conference on Computer Vision and Pattern Recognition Workshop}, 2023, pp. 5548--5558.

\bibitem{tao2024learn}
C.~Tao, C.~Wang, Y.~Zou, X.~Peng, J.~Wu, and J.~Qian, ``Learn suspected anomalies from event prompts for video anomaly detection,'' \emph{arXiv preprint arXiv:2403.01169}, 2024.

\bibitem{wu2024toward}
P.~Wu, J.~Liu, X.~He, Y.~Peng, P.~Wang, and Y.~Zhang, ``Toward video anomaly retrieval from video anomaly detection: new benchmarks and model,'' \emph{IEEE Transactions on Image Processing}, vol.~33, pp. 2213--2225, 2024.

\bibitem{wei2022look}
D.-L. Wei, C.-G. Liu, Y.~Liu, J.~Liu, X.-G. Zhu, and X.-H. Zeng, ``Look, listen and pay more attention: fusing multi-modal information for video violence detection,'' in \emph{Proceedings of the IEEE International Conference on Acoustics, Speech, and Signal Processing}, 2022, pp. 1980--1984.

\bibitem{wei2022msaf}
D.~Wei, Y.~Liu, X.~Zhu, J.~Liu, and X.~Zeng, ``Msaf: multimodal supervise-attention enhanced fusion for video anomaly detection,'' \emph{IEEE Signal Processing Letters}, vol.~29, pp. 2178--2182, 2022.

\bibitem{yu2022modality}
J.~Yu, J.~Liu, Y.~Cheng, R.~Feng, and Y.~Zhang, ``Modality-aware contrastive instance learning with self-distillation for weakly-supervised audio-visual violence detection,'' in \emph{Proceedings of the ACM International Conference on Multimedia}, 2022, pp. 6278--6287.

\bibitem{wu2022weakly}
P.~Wu, X.~Liu, and J.~Liu, ``Weakly supervised audio-visual violence detection,'' \emph{IEEE Transactions on Multimedia}, vol.~25, pp. 1674--1685, 2022.

\bibitem{yuan2024towards}
T.~Yuan, X.~Zhang, K.~Liu, B.~Liu, C.~Chen, J.~Jin, and Z.~Jiao, ``Towards surveillance video-and-language understanding: new dataset baselines and challenges,'' in \emph{Proceedings of the IEEE Conference on Computer Vision and Pattern Recognition}, 2024, pp. 22\,052--22\,061.

\bibitem{zhu2019motion}
Y.~Zhu and S.~Newsam, ``Motion-aware feature for improved video anomaly detection,'' in \emph{Proceedings of the British Machine Vision Conference}, 2019.

\bibitem{zhang2019temporal}
J.~Zhang, L.~Qing, and J.~Miao, ``Temporal convolutional network with complementary inner bag loss for weakly supervised anomaly detection,'' in \emph{Proceedings of the IEEE International Conference on Image Processing}, 2019, pp. 4030--4034.

\bibitem{liu2022collaborative}
Y.~Liu, J.~Liu, M.~Zhao, S.~Li, and L.~Song, ``Collaborative normality learning framework for weakly supervised video anomaly detection,'' \emph{IEEE Transactions on Circuits and Systems II: Express Briefs}, vol.~69, no.~5, pp. 2508--2512, 2022.

\bibitem{feng2021mist}
J.-C. Feng, F.-T. Hong, and W.-S. Zheng, ``Mist: multiple instance self-training framework for video anomaly detection,'' in \emph{Proceedings of the IEEE Conference on Computer Vision and Pattern Recognition}, 2021, pp. 14\,009--14\,018.

\bibitem{zhang2023exploiting}
C.~Zhang, G.~Li, Y.~Qi, S.~Wang, L.~Qing, Q.~Huang, and M.-H. Yang, ``Exploiting completeness and uncertainty of pseudo labels for weakly supervised video anomaly detection,'' in \emph{Proceedings of the IEEE Conference on Computer Vision and Pattern Recognition}, 2023, pp. 16\,271--16\,280.

\bibitem{yang2024text}
Z.~Yang, J.~Liu, and P.~Wu, ``Text prompt with normality guidance for weakly supervised video anomaly detection,'' in \emph{Proceedings of the IEEE Conference on Computer Vision and Pattern Recognition}, 2024, pp. 18\,899--18\,908.

\bibitem{wu2021learning}
P.~Wu and J.~Liu, ``Learning causal temporal relation and feature discrimination for anomaly detection,'' \emph{IEEE Transactions on Image Processing}, vol.~30, pp. 3513--3527, 2021.

\bibitem{tian2021weakly}
Y.~Tian, G.~Pang, Y.~Chen, R.~Singh, J.~W. Verjans, and G.~Carneiro, ``Weakly-supervised video anomaly detection with robust temporal feature magnitude learning,'' in \emph{Proceedings of the IEEE International Conference on Computer Vision}, 2021, pp. 4975--4986.

\bibitem{liu2022decouple}
T.~Liu, C.~Zhang, K.-M. Lam, and J.~Kong, ``Decouple and resolve: transformer-based models for online anomaly detection from weakly labeled videos,'' \emph{IEEE Transactions on Information Forensics and Security}, vol.~18, pp. 15--28, 2022.

\bibitem{chang2021contrastive}
S.~Chang, Y.~Li, S.~Shen, J.~Feng, and Z.~Zhou, ``Contrastive attention for video anomaly detection,'' \emph{IEEE Transactions on Multimedia}, vol.~24, pp. 4067--4076, 2021.

\bibitem{cho2023look}
M.~Cho, M.~Kim, S.~Hwang, C.~Park, K.~Lee, and S.~Lee, ``Look around for anomalies: weakly-supervised anomaly detection via context-motion relational learning,'' in \emph{Proceedings of the IEEE Conference on Computer Vision and Pattern Recognition}, 2023, pp. 12\,137--12\,146.

\bibitem{purwanto2021dance}
D.~Purwanto, Y.-T. Chen, and W.-H. Fang, ``Dance with self-attention: a new look of conditional random fields on anomaly detection in videos,'' in \emph{Proceedings of the IEEE International Conference on Computer Vision}, 2021, pp. 173--183.

\bibitem{huang2022weakly}
C.~Huang, C.~Liu, J.~Wen, L.~Wu, Y.~Xu, Q.~Jiang, and Y.~Wang, ``Weakly supervised video anomaly detection via self-guided temporal discriminative transformer,'' \emph{IEEE Transactions on Cybernetics}, vol.~54, no.~5, pp. 3197--3210, 2022.

\bibitem{zhang2022weakly}
C.~Zhang, G.~Li, Q.~Xu, X.~Zhang, L.~Su, and Q.~Huang, ``Weakly supervised anomaly detection in videos considering the openness of events,'' \emph{IEEE Transactions on Intelligent Transportation Systems}, vol.~23, no.~11, pp. 21\,687--21\,699, 2022.

\bibitem{zhou2023dual}
H.~Zhou, J.~Yu, and W.~Yang, ``Dual memory units with uncertainty regulation for weakly supervised video anomaly detection,'' in \emph{Proceedings of the AAAI Conference on Artificial Intelligence}, vol.~37, no.~3, 2023, pp. 3769--3777.

\bibitem{li2022scale}
G.~Li, G.~Cai, X.~Zeng, and R.~Zhao, ``Scale-aware spatio-temporal relation learning for video anomaly detection,'' in \emph{Proceedings of the European Conference on Computer Vision}, 2022, pp. 333--350.

\bibitem{ye2024learning}
H.~Ye, K.~Xu, X.~Jiang, and T.~Sun, ``Learning spatio-temporal relations with multi-scale integrated perception for video anomaly detection,'' in \emph{Proceedings of the IEEE International Conference on Acoustics, Speech, and Signal Processing}, 2024, pp. 4020--4024.

\bibitem{lin2019social}
S.~Lin, H.~Yang, X.~Tang, T.~Shi, and L.~Chen, ``Social mil: interaction-aware for crowd anomaly detection,'' in \emph{Proceedings of the IEEE International Conference on Advanced Video and Signal-Based Surveillance}, 2019, pp. 1--8.

\bibitem{park2023normality}
S.~Park, H.~Kim, M.~Kim, D.~Kim, and K.~Sohn, ``Normality guided multiple instance learning for weakly supervised video anomaly detection,'' in \emph{Proceedings of the IEEE Winter Conference on Applications of Computer Vision}, 2023, pp. 2665--2674.

\bibitem{lv2023unbiased}
H.~Lv, Z.~Yue, Q.~Sun, B.~Luo, Z.~Cui, and H.~Zhang, ``Unbiased multiple instance learning for weakly supervised video anomaly detection,'' in \emph{Proceedings of the IEEE Conference on Computer Vision and Pattern Recognition}, 2023, pp. 8022--8031.

\bibitem{chen2024prompt}
J.~Chen, L.~Li, L.~Su, Z.-j. Zha, and Q.~Huang, ``Prompt-enhanced multiple instance learning for weakly supervised video anomaly detection,'' in \emph{Proceedings of the IEEE Conference on Computer Vision and Pattern Recognition}, 2024, pp. 18\,319--18\,329.

\bibitem{chen2023mgfn}
Y.~Chen, Z.~Liu, B.~Zhang, W.~Fok, X.~Qi, and Y.-C. Wu, ``Mgfn: magnitude-contrastive glance-and-focus network for weakly-supervised video anomaly detection,'' in \emph{Proceedings of the AAAI Conference on Artificial Intelligence}, vol.~37, no.~1, 2023, pp. 387--395.

\bibitem{gong2022multi}
Y.~Gong, C.~Wang, X.~Dai, S.~Yu, L.~Xiang, and J.~Wu, ``Multi-scale continuity-aware refinement network for weakly supervised video anomaly detection,'' in \emph{Proceedings of the IEEE International Conference on Multimedia and Expo}, 2022, pp. 1--6.

\bibitem{sapkota2022bayesian}
H.~Sapkota and Q.~Yu, ``Bayesian nonparametric submodular video partition for robust anomaly detection,'' in \emph{Proceedings of the IEEE Conference on Computer Vision and Pattern Recognition}, 2022, pp. 3212--3221.

\bibitem{fioresi2023ted}
J.~Fioresi, I.~R. Dave, and M.~Shah, ``Ted-spad: temporal distinctiveness for self-supervised privacy-preservation for video anomaly detection,'' in \emph{Proceedings of the IEEE International Conference on Computer Vision}, 2023, pp. 13\,598--13\,609.

\bibitem{zaheer2023clustering}
M.~Z. Zaheer, A.~Mahmood, M.~Astrid, and S.-I. Lee, ``Clustering aided weakly supervised training to detect anomalous events in surveillance videos,'' \emph{IEEE Transactions on Neural Networks and Learning Systems}, pp. 1--14, 2023.

\bibitem{liu2023distilling}
T.~Liu, K.-M. Lam, and J.~Kong, ``Distilling privileged knowledge for anomalous event detection from weakly labeled videos,'' \emph{IEEE Transactions on Neural Networks and Learning Systems}, pp. 1--15, 2023.

\bibitem{wu2024vadclip}
P.~Wu, X.~Zhou, G.~Pang, L.~Zhou, Q.~Yan, P.~Wang, and Y.~Zhang, ``Vadclip: adapting vision-language models for weakly supervised video anomaly detection,'' in \emph{Proceedings of the AAAI Conference on Artificial Intelligence}, vol.~38, no.~6, 2024, pp. 6074--6082.

\bibitem{wu2024weakly}
P.~Wu, X.~Zhou, G.~Pang, Z.~Yang, Q.~Yan, P.~Wang, and Y.~Zhang, ``Weakly supervised video anomaly detection and localization with spatio-temporal prompts,'' in \emph{Proceedings of the ACM International Conference on Multimedia}, 2024.

\bibitem{joo2023clip}
H.~K. Joo, K.~Vo, K.~Yamazaki, and N.~Le, ``Clip-tsa: clip-assisted temporal self-attention for weakly-supervised video anomaly detection,'' in \emph{Proceedings of the IEEE International Conference on Image Processing}, 2023, pp. 3230--3234.

\bibitem{zhang2024holmes}
H.~Zhang, X.~Xu, X.~Wang, J.~Zuo, C.~Han, X.~Huang, C.~Gao, Y.~Wang, and N.~Sang, ``Holmes-vad: towards unbiased and explainable video anomaly detection via multi-modal llm,'' \emph{arXiv preprint arXiv:2406.12235}, 2024.

\bibitem{lv2024video}
H.~Lv and Q.~Sun, ``Video anomaly detection and explanation via large language models,'' \emph{arXiv preprint arXiv:2401.05702}, 2024.

\bibitem{zanella2024harnessing}
L.~Zanella, W.~Menapace, M.~Mancini, Y.~Wang, and E.~Ricci, ``Harnessing large language models for training-free video anomaly detection,'' in \emph{Proceedings of the IEEE Conference on Computer Vision and Pattern Recognition}, 2024, pp. 18\,527--18\,536.

\bibitem{jeong2023winclip}
J.~Jeong, Y.~Zou, T.~Kim, D.~Zhang, A.~Ravichandran, and O.~Dabeer, ``Winclip: Zero-/few-shot anomaly classification and segmentation,'' in \emph{Proceedings of the IEEE Conference on Computer Vision and Pattern Recognition}, 2023, pp. 19\,606--19\,616.

\bibitem{zhou2024anomalyclip}
Q.~Zhou, G.~Pang, Y.~Tian, S.~He, and J.~Chen, ``Anomalyclip: Object-agnostic prompt learning for zero-shot anomaly detection,'' in \emph{The Twelfth International Conference on Learning Representations}, 2024.

\bibitem{zhu2024toward}
J.~Zhu and G.~Pang, ``Toward generalist anomaly detection via in-context residual learning with few-shot sample prompts,'' in \emph{Proceedings of the IEEE Conference on Computer Vision and Pattern Recognition}, 2024, pp. 17\,826--17\,836.

\bibitem{liu2019exploring}
K.~Liu and H.~Ma, ``Exploring background-bias for anomaly detection in surveillance videos,'' in \emph{Proceedings of the ACM International Conference on Multimedia}, 2019, pp. 1490--1499.

\bibitem{wu2021weakly}
J.~Wu, W.~Zhang, G.~Li, W.~Wu, X.~Tan, Y.~Li, E.~Ding, and L.~Lin, ``Weakly-supervised spatio-temporal anomaly detection in surveillance video,'' in \emph{Proceedings of International Joint Conferences on Artificial Intelligence}, 2021.

\bibitem{dong2016multi}
Z.~Dong, J.~Qin, and Y.~Wang, ``Multi-stream deep networks for person to person violence detection in videos,'' in \emph{Proceedings of the Chinese Conference on Pattern Recognition}, 2016, pp. 517--531.

\bibitem{zhou2017violent}
P.~Zhou, Q.~Ding, H.~Luo, and X.~Hou, ``Violent interaction detection in video based on deep learning,'' in \emph{Journal of Physics: Conference Series}, vol. 844, no.~1, 2017, p. 012044.

\bibitem{peixoto2019toward}
B.~Peixoto, B.~Lavi, J.~P.~P. Martin, S.~Avila, Z.~Dias, and A.~Rocha, ``Toward subjective violence detection in videos,'' in \emph{Proceedings of the IEEE International Conference on Acoustics, Speech, and Signal Processing}, 2019, pp. 8276--8280.

\bibitem{perez2019detection}
M.~Perez, A.~C. Kot, and A.~Rocha, ``Detection of real-world fights in surveillance videos,'' in \emph{Proceedings of the IEEE International Conference on Acoustics, Speech, and Signal Processing}, 2019, pp. 2662--2666.

\bibitem{peixoto2020multimodal}
B.~Peixoto, B.~Lavi, P.~Bestagini, Z.~Dias, and A.~Rocha, ``Multimodal violence detection in videos,'' in \emph{Proceedings of the IEEE International Conference on Acoustics, Speech, and Signal Processing}, 2020, pp. 2957--2961.

\bibitem{sudhakaran2017learning}
S.~Sudhakaran and O.~Lanz, ``Learning to detect violent videos using convolutional long short-term memory,'' in \emph{Proceedings of the IEEE International Conference on Advanced Video and Signal-Based Surveillance}, 2017, pp. 1--6.

\bibitem{hanson2018bidirectional}
A.~Hanson, K.~Pnvr, S.~Krishnagopal, and L.~Davis, ``Bidirectional convolutional lstm for the detection of violence in videos,'' in \emph{Proceedings of the European Conference on Computer Vision Workshop}, 2018, pp. 0--0.

\bibitem{su2020human}
Y.~Su, G.~Lin, J.~Zhu, and Q.~Wu, ``Human interaction learning on 3d skeleton point clouds for video violence recognition,'' in \emph{Proceedings of the European Conference on Computer Vision}, 2020, pp. 74--90.

\bibitem{singh2018eye}
A.~Singh, D.~Patil, and S.~Omkar, ``Eye in the sky: real-time drone surveillance system (dss) for violent individuals identification using scatternet hybrid deep learning network,'' in \emph{Proceedings of the IEEE Conference on Computer Vision and Pattern Recognition Workshop}, 2018, pp. 1629--1637.

\bibitem{cheng2021rwf}
M.~Cheng, K.~Cai, and M.~Li, ``Rwf-2000: an open large scale video database for violence detection,'' in \emph{Proceedings of the International Conference on Pattern Recognition}, 2021, pp. 4183--4190.

\bibitem{shang2022multimodal}
Y.~Shang, X.~Wu, and R.~Liu, ``Multimodal violent video recognition based on mutual distillation,'' in \emph{Proceedings of the Chinese Conference on Pattern Recognition and Computer Vision}, 2022, pp. 623--637.

\bibitem{garcia2023human}
G.~Garcia-Cobo and J.~C. SanMiguel, ``Human skeletons and change detection for efficient violence detection in surveillance videos,'' \emph{Computer Vision and Image Understanding}, vol. 233, p. 103739, 2023.

\bibitem{su2022violence}
J.~Su, P.~Her, E.~Clemens, E.~Yaz, S.~Schneider, and H.~Medeiros, ``Violence detection using 3d convolutional neural networks,'' in \emph{Proceedings of the IEEE International Conference on Advanced Video and Signal-Based Surveillance}, 2022, pp. 1--8.

\bibitem{del2016discriminative}
A.~Del~Giorno, J.~A. Bagnell, and M.~Hebert, ``A discriminative framework for anomaly detection in large videos,'' in \emph{Proceedings of the European Conference on Computer Vision}, 2016, pp. 334--349.

\bibitem{tudor2017unmasking}
R.~Tudor~Ionescu, S.~Smeureanu, B.~Alexe, and M.~Popescu, ``Unmasking the abnormal events in video,'' in \emph{Proceedings of the IEEE International Conference on Computer Vision}, 2017, pp. 2895--2903.

\bibitem{liu2018classifier}
Y.~Liu, C.-L. Li, and B.~P{\'o}czos, ``Classifier two sample test for video anomaly detections,'' in \emph{Proceedings of the British Machine Vision Conference}, 2018, p.~71.

\bibitem{wang2018detecting}
S.~Wang, Y.~Zeng, Q.~Liu, C.~Zhu, E.~Zhu, and J.~Yin, ``Detecting abnormality without knowing normality: a two-stage approach for unsupervised video abnormal event detection,'' in \emph{Proceedings of the ACM International Conference on Multimedia}, 2018, pp. 636--644.

\bibitem{pang2020self}
G.~Pang, C.~Yan, C.~Shen, A.~v.~d. Hengel, and X.~Bai, ``Self-trained deep ordinal regression for end-to-end video anomaly detection,'' in \emph{Proceedings of the IEEE Conference on Computer Vision and Pattern Recognition}, 2020, pp. 12\,173--12\,182.

\bibitem{hu2022detecting}
J.~Hu, G.~Yu, S.~Wang, E.~Zhu, Z.~Cai, and X.~Zhu, ``Detecting anomalous events from unlabeled videos via temporal masked auto-encoding,'' in \emph{Proceedings of the IEEE International Conference on Multimedia and Expo}, 2022, pp. 1--6.

\bibitem{lin2022causal}
X.~Lin, Y.~Chen, G.~Li, and Y.~Yu, ``A causal inference look at unsupervised video anomaly detection,'' in \emph{Proceedings of the AAAI Conference on Artificial Intelligence}, vol.~36, no.~2, 2022, pp. 1620--1629.

\bibitem{yu2022deep}
G.~Yu, S.~Wang, Z.~Cai, X.~Liu, C.~Xu, and C.~Wu, ``Deep anomaly discovery from unlabeled videos via normality advantage and self-paced refinement,'' in \emph{Proceedings of the IEEE Conference on Computer Vision and Pattern Recognition}, 2022, pp. 13\,987--13\,998.

\bibitem{zaheer2022generative}
M.~Z. Zaheer, A.~Mahmood, M.~H. Khan, M.~Segu, F.~Yu, and S.-I. Lee, ``Generative cooperative learning for unsupervised video anomaly detection,'' in \emph{Proceedings of the IEEE Conference on Computer Vision and Pattern Recognition}, 2022, pp. 14\,744--14\,754.

\bibitem{al2024coarse}
A.~Al-lahham, N.~Tastan, M.~Z. Zaheer, and K.~Nandakumar, ``A coarse-to-fine pseudo-labeling (c2fpl) framework for unsupervised video anomaly detection,'' in \emph{Proceedings of the IEEE Winter Conference on Applications of Computer Vision}, 2024, pp. 6793--6802.

\bibitem{he2022masked}
K.~He, X.~Chen, S.~Xie, Y.~Li, P.~Doll{\'a}r, and R.~Girshick, ``Masked autoencoders are scalable vision learners,'' in \emph{Proceedings of the IEEE Conference on Computer Vision and Pattern Recognition}, 2022, pp. 16\,000--16\,009.

\bibitem{li2021deep}
T.~Li, Z.~Wang, S.~Liu, and W.-Y. Lin, ``Deep unsupervised anomaly detection,'' in \emph{Proceedings of the IEEE Winter Conference on Applications of Computer Vision}, 2021, pp. 3636--3645.

\bibitem{liu2019margin}
W.~Liu, W.~Luo, Z.~Li, P.~Zhao, S.~Gao \emph{et~al.}, ``Margin learning embedded prediction for video anomaly detection with a few anomalies,'' in \emph{Proceedings of the International Joint Conference on Artificial Intelligence}, 2019, pp. 3023--3030.

\bibitem{acsintoae2022ubnormal}
A.~Acsintoae, A.~Florescu, M.-I. Georgescu, T.~Mare, P.~Sumedrea, R.~T. Ionescu, F.~S. Khan, and M.~Shah, ``Ubnormal: new benchmark for supervised open-set video anomaly detection,'' in \emph{Proceedings of the IEEE Conference on Computer Vision and Pattern Recognition}, 2022, pp. 20\,143--20\,153.

\bibitem{zhu2022towards}
Y.~Zhu, W.~Bao, and Q.~Yu, ``Towards open set video anomaly detection,'' in \emph{Proceedings of the European Conference on Computer Vision}, 2022, pp. 395--412.

\bibitem{ding2022catching}
C.~Ding, G.~Pang, and C.~Shen, ``Catching both gray and black swans: Open-set supervised anomaly detection,'' in \emph{Proceedings of the IEEE Conference on Computer Vision and Pattern Recognition}, 2022, pp. 7388--7398.

\bibitem{zhu2024anomaly}
J.~Zhu, C.~Ding, Y.~Tian, and G.~Pang, ``Anomaly heterogeneity learning for open-set supervised anomaly detection,'' in \emph{Proceedings of the IEEE Conference on Computer Vision and Pattern Recognition}, 2024, pp. 17\,616--17\,626.

\bibitem{lu2020few}
Y.~Lu, F.~Yu, M.~K.~K. Reddy, and Y.~Wang, ``Few-shot scene-adaptive anomaly detection,'' in \emph{Proceedings of the European Conference on Computer Vision}, 2020, pp. 125--141.

\bibitem{hu2021adaptive}
Y.~Hu, X.~Huang, and X.~Luo, ``Adaptive anomaly detection network for unseen scene without fine-tuning,'' in \emph{Proceedings of the Chinese Conference on Pattern Recognition and Computer Vision}, 2021, pp. 311--323.

\bibitem{huang2022boosting}
X.~Huang, Y.~Hu, X.~Luo, J.~Han, B.~Zhang, and X.~Cao, ``Boosting variational inference with margin learning for few-shot scene-adaptive anomaly detection,'' \emph{IEEE Transactions on Circuits and Systems for Video Technology}, vol.~33, no.~6, pp. 2813--2825, 2022.

\bibitem{aich2023cross}
A.~Aich, K.-C. Peng, and A.~K. Roy-Chowdhury, ``Cross-domain video anomaly detection without target domain adaptation,'' in \emph{Proceedings of the IEEE Winter Conference on Applications of Computer Vision}, 2023, pp. 2579--2591.

\bibitem{li2013anomaly}
W.~Li, V.~Mahadevan, and N.~Vasconcelos, ``Anomaly detection and localization in crowded scenes,'' \emph{IEEE Transactions on Pattern Analysis and Machine Intelligence}, vol.~36, no.~1, pp. 18--32, 2013.

\bibitem{sun2022human}
Z.~Sun, Q.~Ke, H.~Rahmani, M.~Bennamoun, G.~Wang, and J.~Liu, ``Human action recognition from various data modalities: A review,'' \emph{IEEE Transactions on Pattern Analysis and Machine Intelligence}, vol.~45, no.~3, pp. 3200--3225, 2022.

\bibitem{zhu2024vision+}
Y.~Zhu, Y.~Wu, N.~Sebe, and Y.~Yan, ``Vision+x: a survey on multimodal learning in the light of data,'' \emph{IEEE Transactions on Pattern Analysis and Machine Intelligence}, 2024.

\bibitem{gui2024survey}
J.~Gui, T.~Chen, J.~Zhang, Q.~Cao, Z.~Sun, H.~Luo, and D.~Tao, ``A survey on self-supervised learning: algorithms, applications, and future trends,'' \emph{IEEE Transactions on Pattern Analysis and Machine Intelligence}, 2024.

\bibitem{zhou2024class}
D.~Zhou, Q.~Wang, Z.~Qi, H.~Ye, D.~Zhan, and Z.~Liu, ``Class-incremental learning: a survey,'' \emph{IEEE Transactions on Pattern Analysis and Machine Intelligence}, 2024.

\end{thebibliography}
\vfill

\end{document}